\documentclass{article}
\usepackage[bindingoffset=0.2in,
            left=0.7in,
            right=0.7in,
            top=1in,
            bottom=1in,
            footskip=.25in]{geometry}
\usepackage{graphicx} %
\usepackage[numbers,square,sort&compress]{natbib}
\usepackage{multicol}
\usepackage[T1]{fontenc}
\usepackage[utf8]{inputenc}
\usepackage{lmodern}
\usepackage{amsmath,amssymb}
\usepackage{graphicx}
\usepackage{microtype}
\usepackage[hidelinks]{hyperref}
\providecommand{\doi}[1]{doi: \href{https://doi.org/#1}{\nolinkurl{#1}}}
\usepackage{blkarray}
\usepackage{xcolor}

\hypersetup{pdftitle={BioEmu2: Scalable Emulation of Biomolecular Dynamics}}

\numberwithin{equation}{section}

\title{BioEmu2: Scalable Emulation of Biomolecular Dynamics}

\author{\parbox{0.95\textwidth}{\centering Michael Gastegger$^\dagger$, Adam E. Foster$^\dagger$, José Jiménez-Luna$^\dagger$, Tim Hempel$^\dagger$, Sarah Lewis$^\dagger$, Yu~Xie$^\dagger$, Lixin Sun$^\dagger$, Iryna Zaporozhets$^\dagger$, J. Thorben Frank$^\dagger$, Yaoyi Chen$^\dagger$, Michael Plainer$^\dagger$, Ludwig Winkler$^\dagger$, Pedro B. P. S. Reis$^\dagger$, Sven H. Giese$^\dagger$, 
Roberto Sordillo, Victor García Satorras, Eray Inanc, Kacper Kapuśniak, Soojung Yang, Rishwanth Raghu, Rachel Kim, Gabriel Dernbach, Deniz Gunceler, Hannes Schulz, Bonnie Kruft, Marwin Segler, Christopher~M.~Bishop, Cecilia Clementi, Ryota Tomioka, Frank Noé$^*$\\[0.5em]
{$^\dagger$ Equal contribution\\$^*$Correspondence to \texttt{franknoe@microsoft.com}}%
}}
\date{}

\begin{document}

\maketitle

\begin{abstract}
Despite major advances in structure prediction, modelling the motions and interactions that drive biomolecular function at scale remains a central challenge. We introduce BioEmu2, a generative model for atomically resolved biomolecular structures combining static structure prediction, equilibrium sampling and time evolution, that samples orders of magnitude faster than molecular dynamics (MD). Trained on experimental and synthetic structures and extensive MD data, it supports multi-component systems containing proteins, nucleic acids, modified residues, ligands and ions. Structure-prediction accuracy is comparable to top open-weight biomolecular folding models, and a context parallel mode enables prediction of megadalton complexes.
Equilibrium distributions approximate MD references with free-energy errors near 1 kcal/mol across tested
proteins and complexes. Time propagation predicts long-timescale dynamical trajectories and kinetics of binding and conformational changes. Generated structures can be controlled by conditioning on residue contacts or steering with user-defined bias potentials. Enhanced sampling enables free energy calculation involving rare states. BioEmu2 makes the study of biomolecular ensembles and dynamics scalable.
\end{abstract}

\section{Introduction}
\label{sec:introduction}

Proteins, DNA and RNA form the basic building blocks of life, performing functions
from oxygen transport and muscle contraction to storing and processing genetic
information. Their activities combine three
recurring processes: (i) conformational changes, which alter a molecule's
three-dimensional arrangement; (ii) binding and dissociation, which
assemble complexes and capture or release ligands; and (iii) chemical
reactions, which transform substrates or act as molecular switches. These processes are often coupled.
In haemoglobin for example, oxygen binding shifts the conformational
equilibrium from the low-affinity tense (T) state toward the
high-affinity relaxed (R) state. This coupling supports cooperative
oxygen loading in the lungs and unloading at the lower oxygen partial
pressures in tissues \citep{perutz1970haemoglobin}. In adenylate kinase, the catalytic cycle
connects nucleotide binding, domain closure, phosphoryl transfer and
reopening for product release \citep{henzlerwildman2007intrinsic}. These molecular activities form
interconnected pathways that underpin cellular function. A scalable means of understanding
molecular function and its disruption in disease could
transform drug-target selection, clarify mechanisms of action, and
improve predictions of drug efficacy and safety.

Next-generation sequencing has made molecular sequences broadly
accessible \citep{metzker2010sequencing}, while AlphaFold, RoseTTAFold and their all-atom
extensions have transformed the prediction of molecular structures
and complexes \citep{jumper2021alphafold,abramson2024alphafold3,baek2021rosettafold,krishna2024rosettafoldallatom}. A predicted structure alone, however, does not
specify which conformations are populated, how rapidly they
interconvert, or how ligand binding changes these properties.
Understanding function therefore requires information about
structural ensembles and their dynamics.

Biophysical experiments can reveal many aspects of biomolecular function,
but usually suffer from either low throughput or low resolution.
E.g.,~cryo-electron microscopy can resolve different conformational states,
and nuclear magnetic resonance spectroscopy measures structural
fluctuations and exchange processes \citep{bai2015cryoem,kleckner2011nmrdynamics}, but both are expensive and require lengthy
sample preparation, data collection and data analysis. 
A broad range of biophysical techniques probe biomolecular shape,
dynamics and interactions, including fluorescence and infrared
spectroscopy, small-angle X-ray scattering, and hydrogen-deuterium
exchange mass spectrometry \citep{lerner2018smfret,ganim2008amideir,svergun2003saxs,engen2009hdx}. Cell-free expression,
miniaturization and cDNA display expand the scale of functional
characterization, enabling high-throughput measurements of biochemical
activity and protein stability \citep{markin2021htmek,tsuboyama2023megascale}. Even when scaled up, these
approaches measure low-dimensional observables rather than directly
resolving the full atomistic conformational ensemble.

Atomistic molecular dynamics (MD) describes the detailed time-resolved evolution of biomolecular structure subject to a chosen force field.
Its predictive power is limited by the accuracy of the interaction
potential, but especially by the computational cost of adequate
sampling. Folding, conformational rearrangements and
association can involve rare barrier crossings, requiring extensive simulation timescales, as illustrated by simulations of protein folding and protein--protein association \citep{lindorfflarsen2011folding,plattner2017association}. Specialized hardware and
enhanced sampling algorithms extend the accessible timescales, but the computational burden of exploring complex conformational landscapes still remains high \citep{lindorfflarsen2011folding,plattner2017association,henin2022enhancedsampling}.
Machine-learned coarse-grained models reduce this burden by learning
effective interactions for fewer degrees of freedom \citep{wang2019cg,charron2025cg}.
Recent transferable models reproduce structural and
thermodynamic properties of atomistic simulations while achieving
orders-of-magnitude computational savings on benchmark proteins \citep{charron2025cg}.
Yet extending this accuracy across small molecules, recovering atomistic
detail and calibrating kinetics remain challenging.

Generative models learn to directly sample a target distribution rather than resolving every
intervening simulation step. Boltzmann generators sample distributions
defined by known energy functions \citep{noe2019boltzmanngenerators} but have limited capability to scale to large systems. DiG, AlphaFlow and
BioEmu learn molecular ensembles from data \citep{zheng2024dig,jing2024alphaflow,lewis2025bioemu}, but are coarse-grained single-protein models. Timewarp, MDGen and implicit transfer operators extend this approach to time-separated
configurations or trajectory segments \citep{klein2023timewarp,jing2024mdgen,schreiner2023ito}, but neither of these are biomolecular foundation models that can generalize over a wide range of complex structures.

Here we introduce BioEmu2, which unifies three modes in one atomistic framework:
(1) structure prediction, with accuracy comparable to current open-weight biomolecular folding models \citep{wohlwend2024boltz1,zhang2026protenix,opendde,Candido2026ESMFold2} on FoldBench and Runs N' Poses \citep{xu2025benchmarking,vskrinjar2026evaluating}. 
(2) approximate room-temperature equilibrium sampling reproducing free energy surfaces of extensive MD simulations within 1 kcal/mol free energy error and covering the major minima; 
(3) time propagation, which allows structures to be propagated an arbitrary time lag $\tau$ in the future, to emulate long-timescale MD trajectories and predict kinetics of conformational changes and binding. 
BioEmu2 supports multicomponent systems containing proteins, DNA, RNA, modified
residues, ligands and ions at heavy-atom resolution. 
Context parallelism enables inference on systems with over 5000 residues.
Contact conditioning and inference-time steering provide control over
structure generation. Enhanced sampling improves coverage of rare
states, and valid reweighting can recover statistics of the learned
base ensemble.
Several capabilities remain experimental, particularly
slow kinetics, and require larger high-quality experimental datasets
and further validation. By accelerating many MD-style sampling tasks
by orders of magnitude, BioEmu2 enables rapid iteration toward
scalable emulation of biomolecular dynamics and function.

\section{BioEmu2 Model}
\label{sec:bioemu2-model}

\begin{figure}[tp]
    \centering
    \includegraphics[width=1\linewidth]{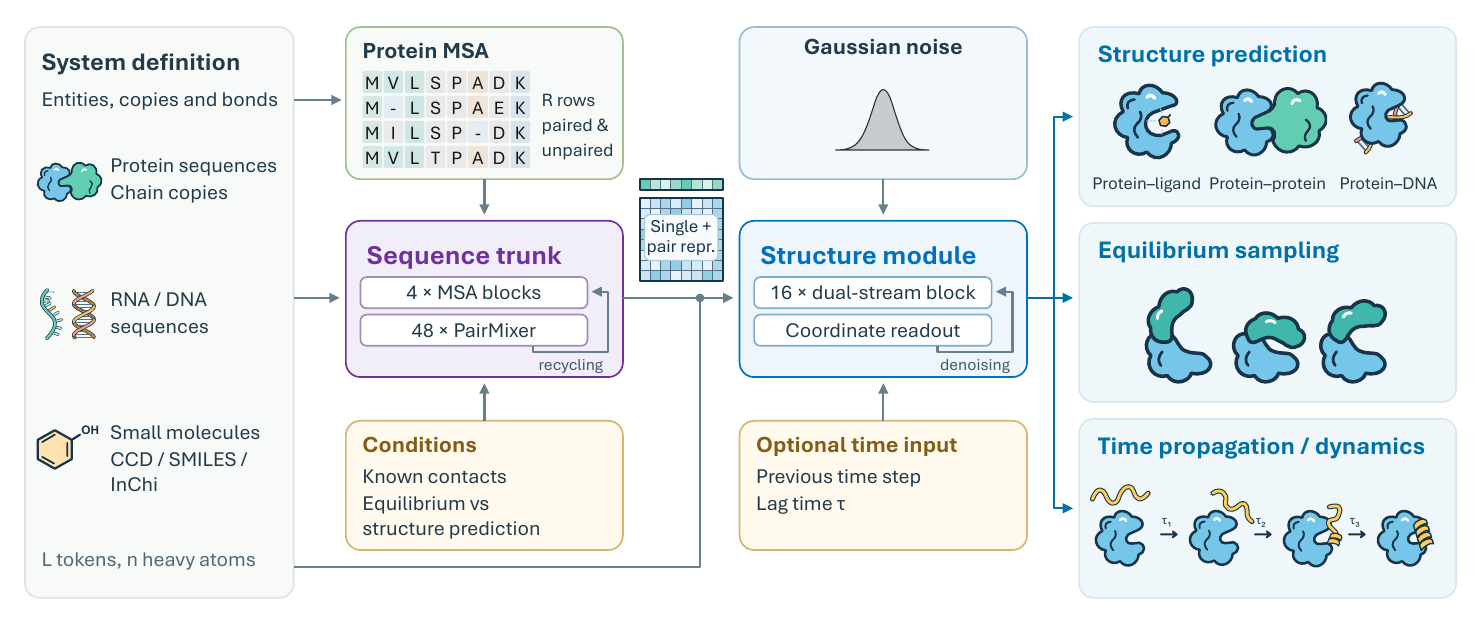}
    \caption{\textbf{BioEmu2 Overview}: High-level architecture of BioEmu2 -- see Sec. \ref{sec:methods} for the detailed architecture of sequence and structure modules. BioEmu2 has three primary modes of usage: static structure prediction, sampling of equilibrium structures and time propagation of structures for emulating molecular dynamics.}
    \label{fig:bioemu2-architecture}
\end{figure}

BioEmu2 is a generative model that can predict biomolecular structures, ensembles and dynamics with atomistic resolution. In contrast to the first version of BioEmu \citep{lewis2025bioemu}, BioEmu2 can model molecular systems consisting of any number of proteins, RNA/DNA, post-translational modifications, lipids, small molecules and ions (Fig.~\ref{fig:bioemu2-architecture}).

At a high level, the BioEmu2 model separates the learning problem into two complementary parts: biological reasoning in the sequence trunk and coordinate generation in the structure module~\citep{jumper2021alphafold,abramson2024alphafold3}. 
In the trunk, a tokenized description of the biomolecular system, $\mathcal{S}$, is obtained by representing protein chains by their amino acid composition, nucleic acids via their nucleotides and other species directly by their individual constituent atoms.
Protein chains are further augmented with evolutionary information in the form of multiple sequence alignments (MSAs) \cite{steinegger2017mmseqs2,mirdita2022colabfold}.
Optional trunk conditioning inputs, $\mathcal{C}_\mathrm{trunk}$, can be provided to enable inference-time control of the structure generation.
These three inputs are then processed by the sequence trunk
\begin{equation}
\begin{aligned}
(\mathcal{S},\mathrm{MSA},\mathcal{C}_\mathrm{trunk})
&\xrightarrow{\text{ sequence trunk }}
\mathcal{E}_\mathrm{trunk}.
\end{aligned}
\end{equation}
BioEmu2 supports two trunk conditioning inputs: a method flag that defines the sample distribution to either aim for precise \textit{structure prediction} or approximate \textit{equilibrium} sampling; and a set (possibly empty) of token-to-token physical contacts that serve as a soft constraint to the model when generating structures.
As a diffusion generative model \cite{sohldickstein2015deep,ho2020denoising,song2021scorebased}, BioEmu2's structure module does not produce structures in a single forward pass, instead it progressively denoises the structure at different noise levels $\sigma$.
Thus a single pass of the structure module takes as input the cached trunk embeddings, along with the coordinates describing a partially noised structure $x_\text{noisy}$ and a noise level $\sigma$
\begin{equation}
\begin{aligned}
(\mathcal{S},\mathcal{E}_\mathrm{trunk}, \mathcal{C}_\textrm{dyn};
x_\text{noisy},\sigma)
&\xrightarrow{\text{ structure module }}
\widehat{x},
\end{aligned}
\end{equation}
yielding the denoised prediction $\widehat{x}$.

Departing from earlier models in the field \citep{abramson2024alphafold3,passaro2025boltz2}, BioEmu2 also emulates dynamics.
In practice, this is achieved through an optional conditioning input to the structure module $\mathcal{C}_\textrm{dyn}=(x_t, \tau)$.
This carries 
the starting structure $x_t$ and a time lag $\tau >0$. 
The model is then trained to sample the distribution $p_\tau(x_{t+\tau} | x_t)$, i.e.~the structural distribution starting in $x_t$ and propagating MD simulation forwards in time over an interval $\tau$.

\paragraph{Inference modes.} The possibility of different conditioning inputs for the sequence trunk and the structure module give raise to three primary combinations of inference modes which are summarized below:

\begin{itemize}
\item \textbf{Structure prediction}
(method = structure prediction,
$\mathcal{C}_{\textrm{dyn}}=\varnothing$):
The structure module generates predicted structures $\widehat{x}$ that mimics static structures obtained by X-ray crystallography or cryo-EM experiments. 

\item \textbf{Equilibrium sampling}
(method = equilibrium, $\mathcal{C}_{\textrm{dyn}}=\varnothing$):
generates conformational samples from a learned approximation to
the equilibrium distribution.

\item \textbf{Time propagation / dynamics}
(method = equilibrium, $\mathcal{C}_{\textrm{dyn}}=(x_t,\tau)$):
Simulates temporal evolution of initial structures $x_t$ for a time lag $\tau$, to time-propagated structures $x  \sim p_\tau(x_{t+\tau} | x_t)$.

\end{itemize}

We note that it is possible to set inputs to the model that combine dynamics conditioning with structure prediction inference mode, but this is not recommended as no training data example uses this. 
MSA computation and trunk embedding are typically more expensive than an individual structure module evaluation. BioEmu2 therefore computes trunk embeddings $\mathcal{E}_{\mathrm{trunk}}$ once and reuses them across many samples or propagation steps.

\section{Results}
We start this section by verifying the competitive performance of BioEmu2 for static structure prediction on established co-folding benchmarks and probe the scalability of our inference to large structures. We then analyse the performance of BioEmu2 for equilibrium sampling before turning to its modelling of biomolecular kinetics. Finally, we highlight how steering and enhanced sampling techniques offer inference-time control over the obtained ensembles and the quality of generated structures. 

\subsection{Structure Prediction}
\paragraph{FoldBench and Runs N' Poses.}
We first assess BioEmu2’s ability to predict accurate individual biomolecular structures.
In structure prediction mode, the model targets structures characteristic of experimental structure data, such as X-ray crystallography and cryo-EM, rather than an equilibrium ensemble.
Using this setting, we compare BioEmu2 to a range of co-folding baseline models on the popular FoldBench benchmark suite~\citep{xu2025benchmarking} and the Runs N' Poses benchmark \citep{vskrinjar2026evaluating} for protein--ligand bound structure prediction.
We selected models for comparison that have a suitable commercial license (excludes AlphaFold3 \cite{abramson2024alphafold3}) and a comparable training data cut-off (excludes Boltz-2 \cite{passaro2025boltz2}), resulting in the following baseline methods: Boltz-1 \cite{wohlwend2024boltz1}, ESMFold2 \cite{Candido2026ESMFold2}, OpenFold3 \cite{openfold3-preview}, Protenix v1 and v2 \cite{zhang2026protenix} and OpenDDE \cite{opendde}.
We do not use confidence heads when aggregating over multiple samples.
The full details of our versions of the two benchmarks are described in Sec.~\ref{sec:method-benchmark-deets}.

\begin{figure}[tp]
    \centering
    \includegraphics[width=1\linewidth]{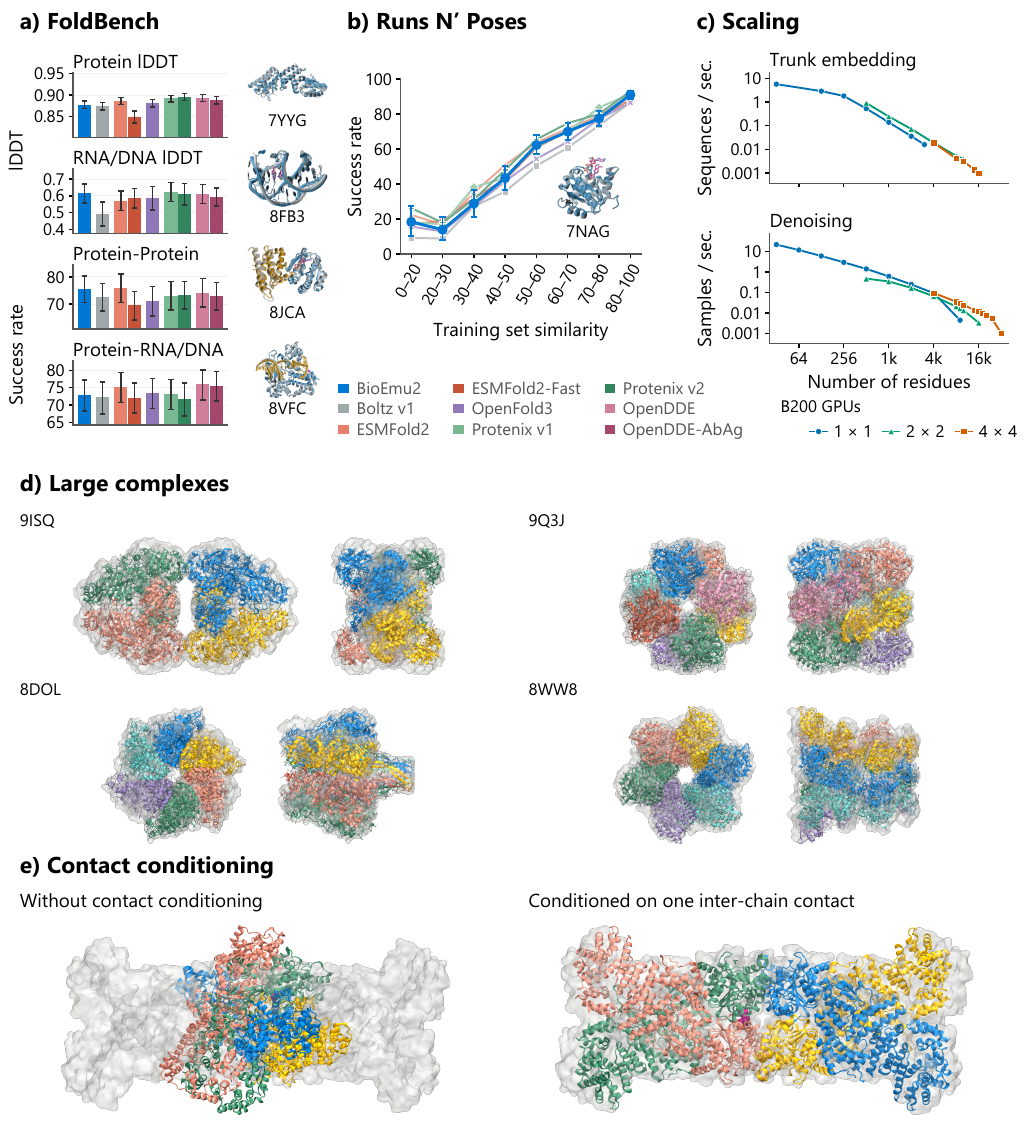}
    \caption{%
    Evaluation of BioEmu2's structure prediction mode and comparison to top open-weight models trained with comparable  cut-off date. 
    Illustrative predicted structures in colour, PDB ground truth in grey. \textbf{a)} Results on FoldBench, grouped by protein, RNA/DNA, protein complexes and protein--RNA/DNA complexes. Legend as in b). \textbf{b)} Runs N' Poses benchmark of high-resolution protein--ligand complexes. \textbf{c)} Computational cost of computing the sequence embedding and sampling structures with given embedding. \textbf{d)} Examples of large complexes computed with context-parallelization on multiple GPUs. \textbf{e)} Illustration of how contact conditioning on a single residue (Asn202 from chains C and D) can fix a failed protein complex structure prediction for 8WY8.}
    \label{fig:structure}
\end{figure}

Across both benchmarks, BioEmu2 achieves accuracy competitive with the evaluated co-folding baselines, establishing structure prediction as a core capability of our model (Fig.~\ref{fig:structure}a,b).
Specifically, in the FoldBench results we see that BioEmu2 achieves competitive lDDT \cite{mariani2013lddt} scores for protein and nucleic acid monomers, and attains strong DockQ-based \cite{basu2016dockq} success rates for chain-to-chain binding. The higher score for protein--protein binding is likely driven by the training data composition, which emphasized protein--protein interaction.
The protein--ligand evaluation from Runs N’ Poses also highlights the ability of BioEmu2 to model ligand placement alongside the surrounding protein structure~\citep{vskrinjar2026evaluating} at a performance level comparable to other strong models.
In the comparison presented in Section~\ref{sec:bioemu2-03-conditioning-and-steering-steering-and-enhanced-sampling}, physicality-steered sampling largely preserves structural prediction accuracy, with modest gains in some categories (Fig.~\ref{fig:steering}b).

\paragraph{Scaling inference to large complexes.}
BioEmu2's structural prediction capability extends to larger molecular assemblies.
To make larger biomolecular systems accessible at inference time, BioEmu2 implements context parallelism (CP) to distribute pair representations across multiple GPUs~\citep{lin2026fold}.
Benchmarks on NVIDIA B200 GPUs demonstrate scaling from a single GPU to 4 GPUs ($2\times2$) and 16 GPUs ($4\times4$) configurations, supporting systems of up to approximately 16,000 residues.
Figure~\ref{fig:structure}c shows that our CP code substantially extends the maximum accessible system size with manageable communication overhead.
At this size, the physicality of the prediction becomes more of a bottleneck than computational feasibility.

BioEmu2 preserves both local structural accuracy and overall assembly organization for a wide range of large protein complexes containing thousands of residues.
Examples include four multimeric complexes in the test set: 9ISQ (3,508 residues), 8DOL (3,558 residues), 9Q3J (4,040 residues) and 8WW8 (5,832 residues), achieving backbone lDDT scores of 0.81, 0.87, 0.94, and 0.80, global RMSDs of 5.5Å, 5.7 Å, 2.0Å and 6.6 Å respectively (Fig.~\ref{fig:structure}d).

The probability to get successful structure prediction out of the box tends to decrease for increasingly large complexes---both because the structure prediction problem is harder, and because it is often not well-defined as large complexes tend to be assembled in the cell rather than forming in solution. In this case, providing a few additional hints such as residue--residue contacts obtained from experimental knowledge can be sufficient to substantially improve BioEmu2's predictions.
As an example, for the 4,020-residue complex 8WY8, specifying just a single inter-chain contact dramatically improves prediction accuracy, reducing RMSD from 71.2 Å to 5.0 Å and increasing DockQ from 0.10 to 0.46 (Fig.~\ref{fig:structure}e).

\subsection{Equilibrium Sampling}
\begin{figure}[tp]
    \centering
    \includegraphics[width=1\linewidth]{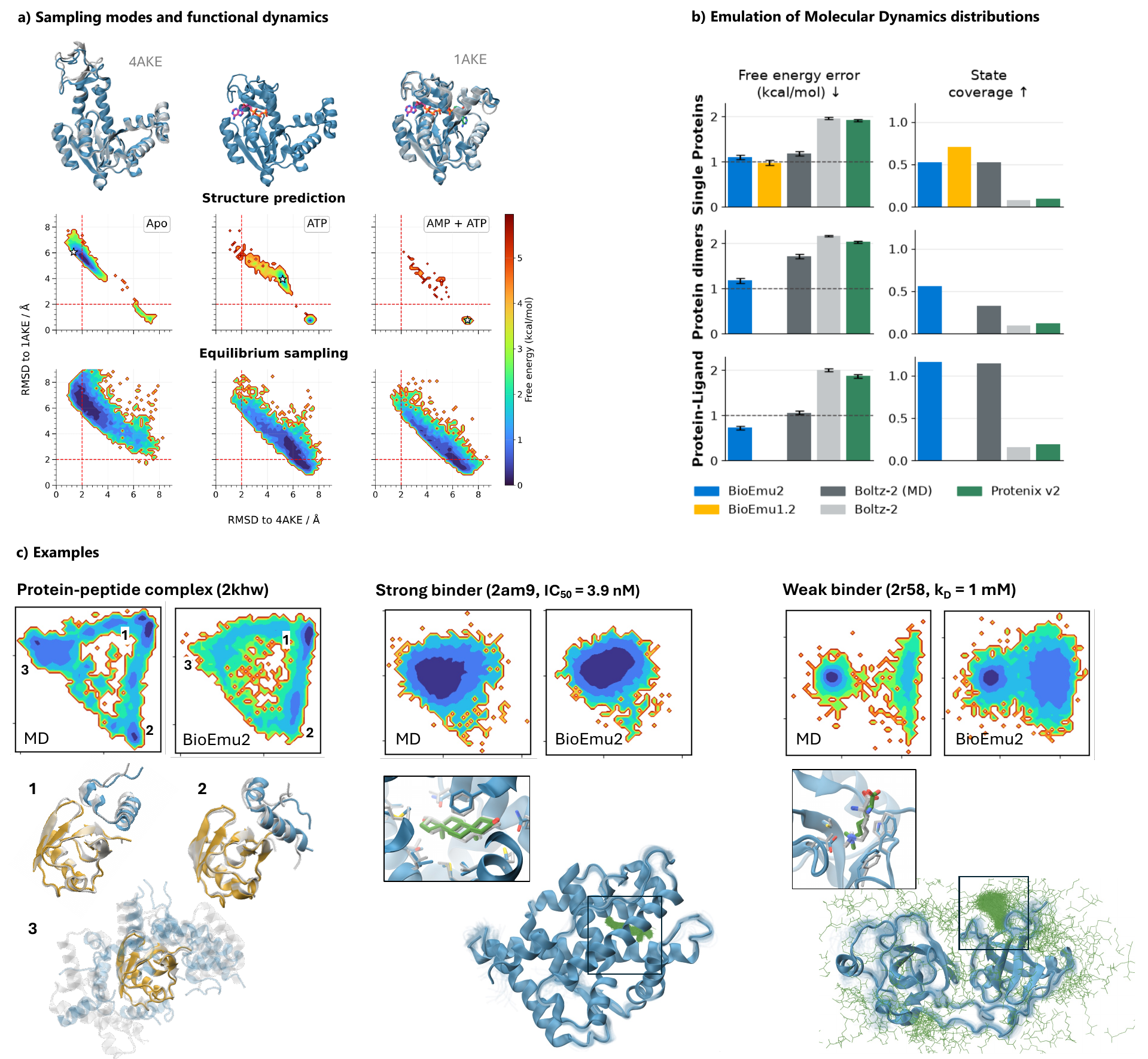}
    \caption{\textbf{Functional dynamics and molecular dynamics emulation.} \textbf{a)} Conformational sampling of adenylate kinase across open (4AKE) and closed (1AKE) states under apo, ATP-, and AMP+ATP-bound conditions. Projections of RMSD to 1AKE versus 4AKE illustrate the difference between standard structure prediction (middle) and equilibrium sampling (bottom) modes. 
  \textbf{b)} Emulation performance of molecular dynamics distributions across single proteins, protein dimers, and protein--ligand complexes. Methods (BioEmu2, BioEmu1.2, Boltz-2 (MD), Boltz-2, and Protenix~v2) are evaluated on free-energy error ($\mathrm{kcal\,mol^{-1}}$, lower is better) and conformational state coverage (higher is better). 
  \textbf{c)} Representative examples comparing MD and BioEmu2 free-energy landscapes alongside corresponding structural ensembles: a multi-state protein--peptide complex (PDB 2KHW, states 1--3), a strong binder (PDB 2AM9, $\mathrm{IC}_{50} = 3.9\,\mathrm{nM}$), and a weak binder exhibiting conformational flexibility (PDB 2R58, $K_{\mathrm{d}} = 1\,\mathrm{mM}$).}
    \label{fig:distributions}
\end{figure}

In equilibrium mode, BioEmu2 generates independent draws from a learned structural distribution for the specified system. Instead of targeting sharp single-state predictions, this mode aims at reproducing the distribution of structures in acqueous solvent at ambient conditions (room temperature and atmospheric pressure). 
This typically leads to higher flexibility and more shallow energy minima than in structure prediction mode (Fig~\ref{fig:distributions}a). 

BioEmu2 is a \textit{Boltzmann emulator}: unlike Boltzmann Generators \citep{noe2019boltzmanngenerators},
which sample from a distribution $p_{eq}(x)$ specified by a known force
field, it learns to sample an approximation to $p_\theta(x)$ from data,
generalizing across proteins and biomolecular systems. This follows
earlier data-driven ensemble models, including AlphaFlow \citep{jing2024alphaflow} and BioEmu \citep{lewis2025bioemu}. Formally, %
\begin{equation} 
x \sim p_\theta(x | \mathcal{S}) \qquad p_\theta(x | \mathcal{S}) \approx p_{eq}(x | \mathcal{S}) 
\label{eq:high-level-bioemu} 
\end{equation} 
where $\mathcal{S}$ is the molecular system, $x$ are the 3D coordinates of all heavy atoms, and $p_\theta$ is BioEmu2's equilibrium distribution aiming to approximate the ground truth equilibrium distribution $p_{eq}$. 
$p_\theta$ is learned from the structure distribution generated by explicit-solvent all-atom MD simulations at 300K temperature and $1$ atm pressure in the isothermal-isobaric ($NpT$) ensemble using Amber force fields \citep{wang2004development}. Solvent and simulation volume are marginalized out, so solvent effects are implicitly learned by the model. 

\paragraph{Free energy surfaces.}
To compare BioEmu2's $p_\theta$ distribution with the  MD data ground truth distribution $p_{eq}$, we first project samples from both distributions onto the two collective variables that govern the slowest transition processes of each test system, computed by time-lagged independent component analysis (TICA)~\cite{perezhernandez2013tica,schwantes2013tica}. This projection resolves the metastable (long-lived) states which (i) have potentially biological relevance, e.g.~as conformations selected by binding partners, and (ii) represent a more challenging and informative test when comparing $p_\theta$ and $p_{eq}$ rather than fast or local fluctuation metrics \cite{olsson2026streetlight}. We compute and compare free energy surfaces $U(x) = -k_BT \ln p_Y(x)$ where $p_Y(x)$ is the 2D-TICA density from either MD or BioEmu2 (Fig.~\ref{fig:distributions})

\paragraph{Free energy error} measures how accurately the model reproduces the relative populations of metastable states. Let $q^{eq}_n$ and $q_n^\theta$ denote the relative populations of metastable state $n$ in the reference and model distributions, respectively, obtained by integrating $p_{eq}$ and $p_\theta$ over the corresponding region of configuration space and normalizing. We subsequently report the mean absolute error over the $M$ metastable states found in the TICA projection: 
\begin{equation} 
\mathrm{MAE}_F = \frac{k_B T}{M}\sum_n \left| \ln q_n^\theta - \ln q^{eq}_n \right|. 
\end{equation}
\paragraph{State coverage} instead measures how much of the reference conformational space is sampled by the model, normalized for sample count. We finely discretize the two-dimensional TICA space into states and report
\begin{equation} 
\mathrm{coverage}
= \frac{|C_\theta \cap C_{\mathrm{ref}}|}
       {\widetilde{c}_{\mathrm{ref}}(N)},
\end{equation} 
where $C_\theta$ and $C_{\mathrm{ref}}$ are the model-visited and reference-visited state sets, and $\widetilde{c}_{\mathrm{ref}}(N)$ is the expected number of distinct states visited by an equally sized sample drawn independently from the reference distribution. Coverage equals one when the model matches the expected reference coverage.

Together, these metrics capture whether BioEmu2 both finds the relevant conformational states and reproduces their relative populations.

\paragraph{Sampling modes and functional dynamics.} 
BioEmu2 allows to choose between standard structure prediction and equilibrium sampling. We probe the effect of the two modes on the example of adenylate kinase, which exhibits a large conformational change upon substrate binding. Using BioEmu2 in structure prediction and equilibrium sampling mode, we generate conformations for the Apo (PDB structure 4AKE), the ATP-bound and the AMP+ATP-bound (PDB structure 1AKE) state (Fig.~\ref{fig:distributions}a). In both modes, BioEmu2 samples structures close to the apo and the holo state, where the relative weight between sampled conformations shifts notably when adding ATP and AMP+ATP to the system, which is in line with experimentally measured PDB structures (stars). By switching between the modes, BioEmu2 transitions from narrowly distributed structure prediction to a continuously distributed ensemble of states, which captures the nature of thermal fluctuations in a biomolecular system. We run the subsequent experiments in equilibrium sampling mode and quantitatively assess the quality of the ensembles generated by BioEmu2.

\paragraph{Emulation of molecular dynamics.}
We now evaluate how BioEmu2 performs on a set of test systems representing different important biomolecular modalities: single proteins, protein dimers, and protein--small molecule systems. 
BioEmu2 samples around 50\% of states found by MD for proteins and protein complexes, and on average over 100\%, i.e.~more states than sampled by and equally sized sample of MD frames for protein--ligand systems (Fig. \ref{fig:distributions}b).
Amongst sampled states, the free energy error is in the range of 0.7 to 1.2 kcal/mol (Fig. \ref{fig:distributions}b).
BioEmu2 strongly outperforms models that have not been trained on MD data in terms of both coverage and free energy error (Proteinx v2 and Boltz-2 default mode shown 
as examples). BioEmu1 is still best on single protein dynamics, likely because it has been fine-tuned on experimental protein stabilities which is not yet the case for the current BioEmu2 release. However, BioEmu1 is a single protein model and therefore unable to compete in the protein--protein and protein--ligand categories. Boltz-2 has a MD mode and in
that mode achieves close to BioEmu2 performance on state coverage, but BioEmu2 as significantly lower free energy error.

Fig~\ref{fig:distributions}c shows TICA landscapes and structures for three concrete example systems and compares them to the MD reference. BioEmu2 consistently matches the TICA landscapes, finding all modes and reasonably approximating their relative proportions, though often with slightly lower energy barriers between metastable states. In the example of a protein--peptide system, we find both the two peptide binding modes as well as an unbound mode, all three of which are also found in the MD reference. Furthermore, for randomly selected examples of a strong and a weak small-molecule binder, BioEmu2 qualitatively matches expectations with regard to the structural ensemble: for the strong binder, all samples are in the bound state and match the crystal structure (2AM9). For the weak binder, both bound and unbound structures are sampled, with the ligand diffusing over the protein surface in the unbound state.

\subsection{Dynamics and Kinetics}
Molecular kinetics can be described by the transition density $p_\tau(y\mid x)$, the probability density of observing configuration $y$ a physical time $\tau$ after configuration $x$. It defines a transfer operator that propagates ensembles of configurations forward in time.
Recent transferable generative models such as Timewarp \citep{klein2023timewarp}, MDGen \citep{jing2024mdgen} and implicit transfer operators (ITO) \citep{schreiner2023ito,vigueradiez2026tito} implicitly learn this density and can sample future configurations directly. BioEmu2 follows this approach and samples
\begin{equation}
   x_{t+\tau}\sim p_{\theta,\tau}(\,\cdot\mid x_t),
\end{equation}
where $p_{\theta,\tau}$ is learned from pairs of configurations $(x_t,x_{t+\tau})$ that are a time lag $\tau$ apart in MD trajectories. The model generates the endpoint $x_{t+\tau}$ without resolving the motion in between, and iterating this step yields trajectories with time steps many orders of magnitude longer than the femtosecond integration step of MD.

For ergodic physical dynamics, the transition density $p_{\tau}(y\mid x)$ loses memory of the starting configuration $x$ at sufficiently long lags $\tau$ and approaches the equilibrium distribution $p_{eq}(y)$. With a suitable parametrization (Sec.~\ref{sec:structure-module-and-sampling-time-propagation-with-an-implicit-transfer-operator}), BioEmu2 reproduces this limit and thereby unifies its transfer operator and its equilibrium distribution in one model.
As the timescales of $\tau$ increase, time propagation thus turns continuously into equilibrium sampling within the same network, and loses all memory of the starting configuration for $\tau \gg 1\,\mathrm{ms}$.
Despite this asymptotic limit, predictions of slower binding, dissociation,
folding, unfolding and other rearrangements are not established as
quantitatively accurate kinetics in the current BioEmu2 release.

\begin{figure}[tp]
    \centering
    \includegraphics[width=1\linewidth]{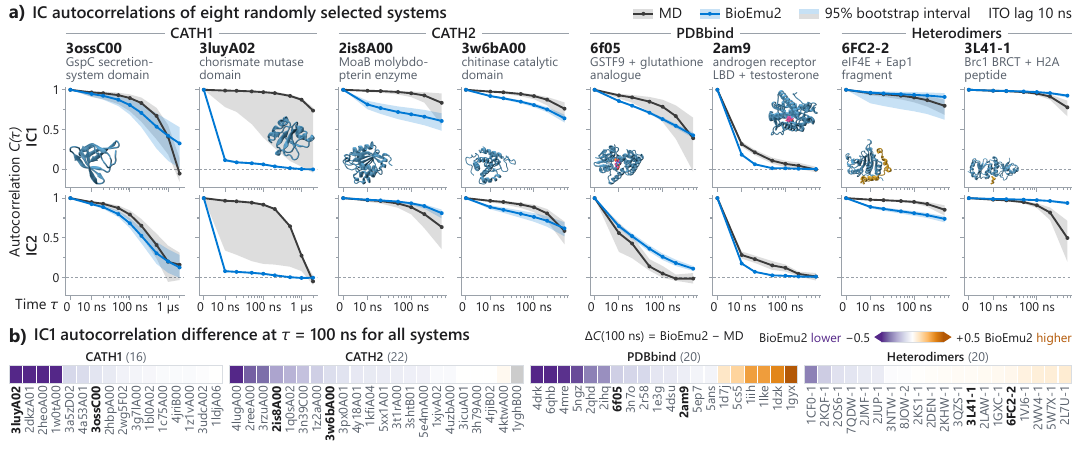}
    \caption{\textbf{Comparing the autocorrelation, $C(\tau)$, of the slowest TICA modes between sequential BioEmu2 ITO rollouts and MD.} 
    \textbf{a)}~$C(\tau)$ of MD and BioEmu2 (95\% bands) for internal coordinates IC1 (top) and IC2 (bottom) of eight systems drawn at random from CATH
   domains, PDBbind protein--ligand complexes and heterodimers.
   \textbf{b)}~$\Delta C$, the autocorrelation of BioEmu2 minus that of MD, of IC1 at $\tau = 100$~ns for 78 systems: $\Delta C < 0$ indicates faster decay in BioEmu2 than in MD, and $\Delta C > 0$ slower decay.
    }
    \label{fig:dynamics-autocorrelation}
\end{figure}

\paragraph{Autocorrelations on MD timescales.} 
Matching equilibrium distributions does not necessarily imply realistic dynamics, because different dynamics can yield the same equilibrium. A direct test for dynamics is whether BioEmu2 trajectories decorrelate as quickly as MD along the slowest motions that MD reveals. To test this, we compare BioEmu2 ITO trajectories with MD for 78 systems: CATH domains, PDBbind protein--ligand complexes and protein heterodimers (Fig.~\ref{fig:dynamics-autocorrelation}). For each system, we project the MD trajectories and 100 BioEmu2 trajectories 
onto the two slowest TICA coordinates of MD, IC1 and IC2. The autocorrelation $C(\tau)$ decays to zero as the trajectories lose memory of earlier configurations, and a faster decay means faster dynamics along that coordinate.
For most systems, the difference from MD at $\tau = 100\,\mathrm{ns}$ lies within the range of differences between random halves of the MD data; where it is larger, BioEmu2 mostly decays faster (Fig.~\ref{fig:dynamics-autocorrelation}). The differences are similar whether $100\,\mathrm{ns}$ is covered by one step of $\tau = 100\,\mathrm{ns}$ or by 100 steps of $\tau = 1\,\mathrm{ns}$, with no clear sign that errors accumulate over many steps. Generally, we observe that BioEmu2 comes close to MD for many systems, but sometimes relaxes too quickly.

\begin{figure}[tp]
    \centering
    \includegraphics[width=1\linewidth]{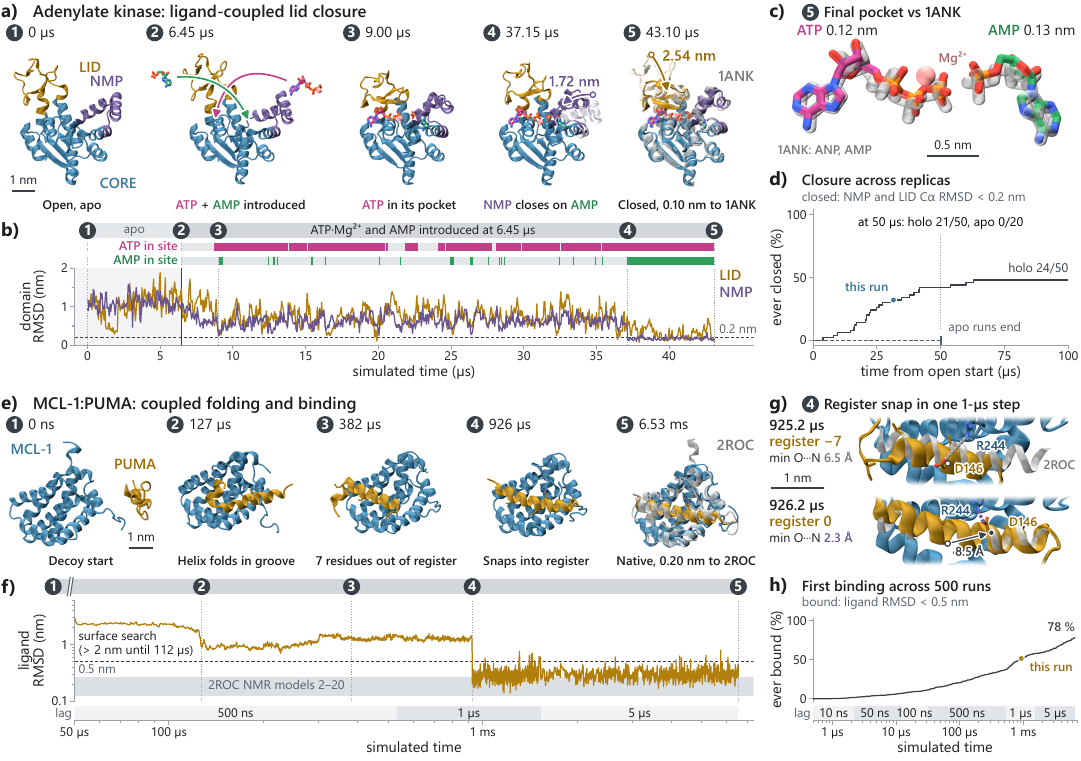}
    \caption{\textbf{BioEmu2 ITO simulations capture ligand-coupled domain closure of adenylate kinase and the coupled folding and binding of PUMA to MCL-1.} Snapshots are samples of single trajectories generated with sequential ITO simulation; numbered events (1--5) mark samples on simulated time. \\
    \textbf{Adenylate kinase (ADK)}.
    \textbf{a)}~Snapshots from one adenylate kinase trajectory before and after ATP$\cdot$Mg$^{2+}$ and AMP addition (50~ns lag). The final pose has a 0.10~nm C$\alpha$ RMSD from the crystal structure 1ANK.
    \textbf{b)}~LID and NMP C$\alpha$ RMSDs to 1ANK after CORE alignment; strips mark ligand poses within 0.2~nm RMSD of their 1ANK counterparts.
    \textbf{c)}~Final binding pocket superposed on 1ANK (ANP and AMP).
    \textbf{d)}~Cumulative fraction of independent trajectories that have closed.\\
    \textbf{MCL-1:PUMA.}
    \textbf{e)}~Snapshots from one PUMA--MCL-1 trajectory (lag increasing from 1~ns to 5~$\mu$s); the endpoint RMSD is to the closest 2ROC NMR model.
   \textbf{f)}~C$\alpha$ RMSD of PUMA to 2ROC (model~1) after alignment on the MCL-1 binding site, with the spread of NMR models 2--20 (band), the binding cutoff of 0.5~nm (dashed) and the ITO lag (strip).
   \textbf{g)}~PUMA D146 and MCL-1 R244 before and after the register shift (4), superposed on 2ROC (snapshots 1~$\mu$s apart).
   \textbf{h)}~Cumulative fraction of independent trajectories that have bound (PUMA C$\alpha$ RMSD below 0.5~nm): 262 of 500 within 1~ms and 391 of 500 within 6.53~ms.
    }
    \label{fig:dynamics-adk-puma}
\end{figure}

\paragraph{Ligand-coupled domain closure of adenylate kinase.} 
We emulate MD simulations with BioEmu2 by generating sequential rollouts ($x_0$, $x_{\tau_1}$, $x_{\tau_1+\tau_2}$, $\dots$, $x_{T}$). Here one can choose the lag times equally, $\tau_1=\tau_2=\cdots$, or differently, e.g.~to finely resolve initial relaxation processes and make long time steps later to cover a large timescale $T$. 
We first illustrate Adenylate kinase (ADK), a classic example of ligand binding coupled to large domain motions: its LID and NMP domains close over ATP and AMP to form the catalytically active state \citep{berry1994adk,henzlerwildman2007intrinsic}. Equilibrium sampling with BioEmu2 captures the ligand-dependent states of ADK (Fig.~\ref{fig:distributions}a) but not the order in which the ligands bind and the domains close. In the trajectory shown, at a lag of $50\,\mathrm{ns}$ (Fig.~\ref{fig:dynamics-adk-puma}a--c), we place ATP$\cdot$Mg$^{2+}$ and AMP beside the open apo enzyme, each closer to the binding site of the other. Within $3\,\mu\mathrm{s}$, both ligands find their own sites and then ATP stays mostly bound, whereas the NMP domain takes almost $30\,\mu\mathrm{s}$ until it closes over AMP, followed by the LID less than $1\,\mu\mathrm{s}$ later. At $43\,\mu\mathrm{s}$, the enzyme is within $0.1\,\mathrm{nm}$ C$\alpha$ RMSD of the closed crystal structure 1ANK \citep{berry1994adk}, with both ligands in their crystallographic poses (Fig.~\ref{fig:dynamics-adk-puma}c). Across 50 trajectories, about 40\% close within $50\,\mu\mathrm{s}$ when the ligands are present, but none when ligands are not present (Fig.~\ref{fig:dynamics-adk-puma}d). BioEmu2 thus propagates the enzyme together with its ligands and Mg$^{2+}$ through coupled binding and domain closure, and it provides testable predictions about the mechanism of ligand-induced conformational transitions.

\paragraph{Coupled folding and binding of PUMA to MCL-1.} 
Many intrinsically disordered proteins fold only when they bind with a partner, and the routes by which they reach their bound structure are difficult to observe directly. The BH3 domain of PUMA is disordered on its own and folds into a helix as it binds the groove of the protein MCL-1, which makes it a model system for coupled folding and binding \citep{day20082roc,rogers2013puma,charron2025cg}. We investigate if BioEmu2 finds the native complex when PUMA starts outside the groove, and by which route. We run 500 trajectories of a total of $T=6.5\,\mathrm{ms}$ each from one starting structure, with a lag that grows from $1\,\mathrm{ns}$ to $5\,\mu\mathrm{s}$, adding up to more than $3\,\mathrm{s}$ of simulated time. In the trajectory shown (Fig.~\ref{fig:dynamics-adk-puma}e--g), PUMA explores the surface of MCL-1 for about $100\,\mu\mathrm{s}$ and then folds into a helix in the groove, which settles seven residues (about two helical turns) out of register. This misregistered helix persists for more than half a millisecond until, within a single $1\,\mu\mathrm{s}$ step, it snaps into the native register, and D146 of PUMA forms a salt bridge. PUMA then stays in the native register, ending close to the NMR structure 2ROC \citep{day20082roc}. Of 500 trajectories, 78\% bind (Fig.~\ref{fig:dynamics-adk-puma}h), and about three quarters of these first pass through a non-native helical intermediate state.


\subsection{Steering and Enhanced Sampling}
\label{sec:bioemu2-03-conditioning-and-steering-steering-and-enhanced-sampling}

BioEmu2 sample generation can be manipulated at inference time using steering.
This can be used both to enhance physical quality of generated samples and to improve rare state coverage using enhanced sampling techniques \citep{xie2026enhanceddiffusion}.
Let $p_\theta(x)$ denote the unsteered model distribution, a potential funtion $v(x)$
defines a desired tilted distribution,
\begin{equation}
p_v(x)\propto p_{\theta}(x)\exp[-v(x)].
\end{equation}
Potentials can penalize geometric defects, favour contacts or reference
conformations, or target a range of a collective variable. 
For physicality steering, we use guidance on detached denoised coordinates with optional Feynman-Kac resampling \cite{wohlwend2024boltz1,singhal2025fk}. 
For the enhanced sampling, we differentiate the potential through the denoiser and use a discrete transition-ratio weighting scheme adapted from the Twisted Diffusion Sampler (TDS) \citep{wu2023practical}.

\begin{figure}[tp]
    \centering
    \includegraphics[width=1\linewidth]{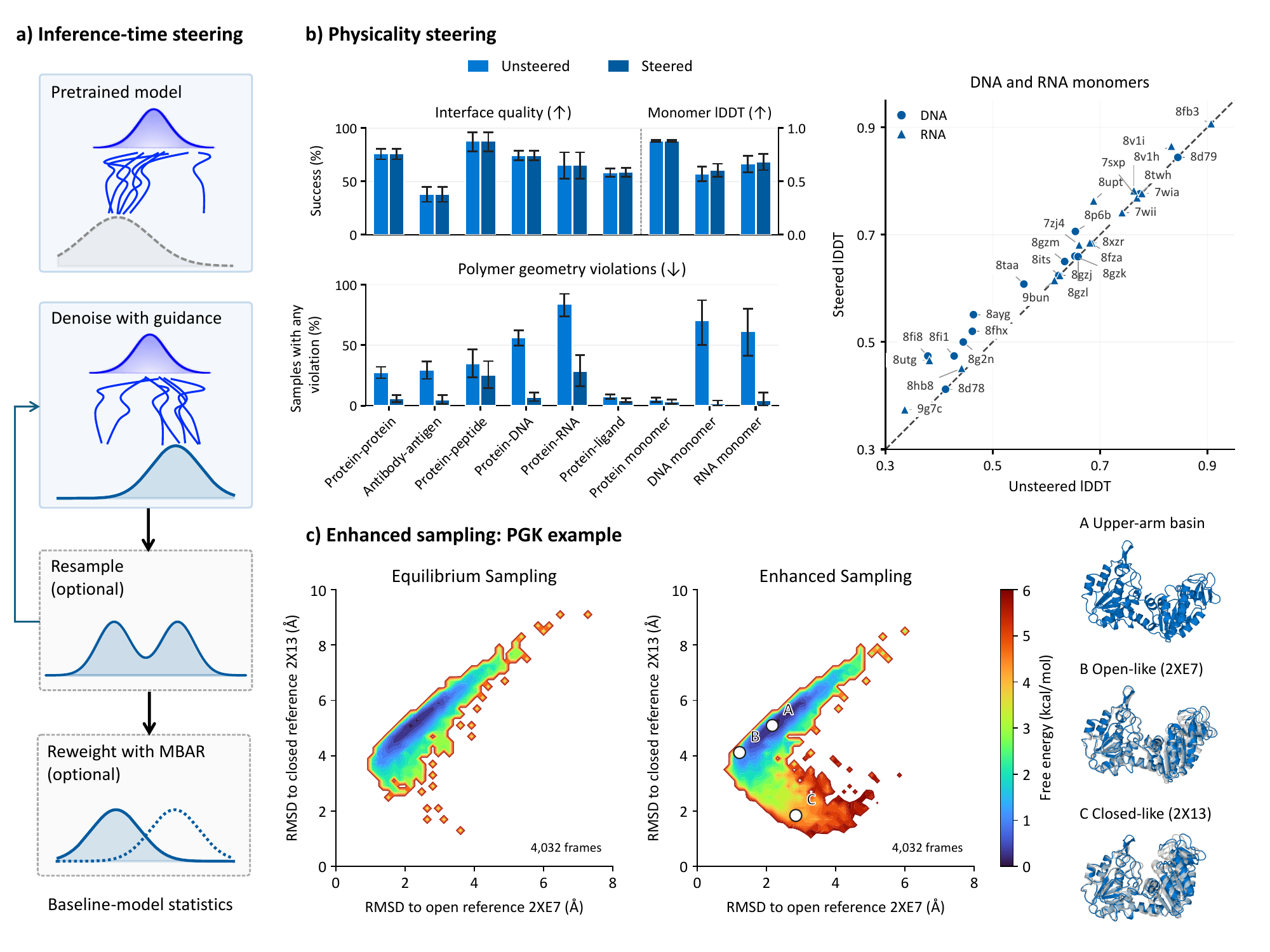}
    \caption{
    \textbf{Inference-time steering for polymer geometry and enhanced conformational sampling.} 
\textbf{a)} Schematic of inference-time steering. Includes guidance, optional particle resampling, and optional post-sampling reweighting to estimate baseline-model statistics.
\textbf{b)} Unsteered (light blue) and steered (dark blue) FoldBench results. Upper bars show best-of-five interface success and mean oracle monomer lDDT; lower bars show the fraction of all samples with any OST polymer clash, bond-length, or bond-angle violation. Error bars: 95\% bootstrap intervals. The scatter compares independently selected oracle lDDT for 14 DNA (circles) and 15 RNA (triangles) monomers.
\textbf{c)} Apo PGK: unsteered sampling and an MBAR-reweighted zero-bias estimate incorporating TDS-biased samples. Coordinates are C$\alpha$ RMSDs to open 2XE7 and closed 2X13. Colors show free energies at 300 K in kcal/mol. A-C mark upper-arm, open-like, and closed-like conformations; predictions are blue and PDB references gray. Sampling, alignment, and reweighting details are in Sec.~\ref{sec:steering-and-enhanced-sampling}.
}
    \label{fig:steering}
\end{figure}

\paragraph{Physicality steering.} 
For structure prediction, we use geometric restraints to discourage bond distortions, steric overlaps, and stereochemical defects without imposing a particular target fold. 
Our implementation combines potentials derived from Boltz \cite{wohlwend2024boltz1,passaro2025boltz2} with the addition of explicit polymer geometry and clash terms, including restraints for nucleic acids (see Sec.~\ref{sec:steering-and-enhanced-sampling} for details). 
The results in Fig.~\ref{fig:steering}b compare unsteered sampling with physicality steering applied during approximately the final 20\% of denoising across the FoldBench benchmark. 
For most categories structural accuracy benchmarks are unaffected by steering, however notable improvement of mean best-of-five lDDT is seen for DNA monomers (0.571 to 0.604) and RNA monomers (0.660 to 0.679). For individual DNA/RNA systems the improvement can be substantial (Fig.~\ref{fig:steering}b, right).
Across categories, significantly fewer steered samples had a polymer clash, bond-length outlier, or bond-angle outlier (Fig.~\ref{fig:steering}b, bottom). 
For DNA and RNA monomers, the unphysical fraction decreases from 70.0\% to 1.4\% and from 61.3\% to 4.0\%, respectively. 

\paragraph{Enhanced diffusion sampling for rare states.} 
Enhanced sampling is a family of methods used in MD simulation to enable free energy calculations with large
free energy differences \cite{zwanzig1954perturbation,torrie1977umbrella}. The principle is to artificially increase the frequency of states that are rare in equilibrium using a bias potential, and to then reweight the biased samples so as to recover unbiased free energy differences.
We apply the same principle to diffusion model sampling \cite{xie2026enhanceddiffusion}: inference-time steering methods sush as TDS \cite{wu2023practical} or Feynman-Kac Corrector (FKC) \cite{skreta2025fkc}  generate biased ensembles, 
and the multistate Bennett acceptance ratio (MBAR) \cite{bartels2000biased,kong2003theory,shirts2008mbar} combines them to estimate the zero-bias model distribution.

We use phosphoglycerate kinase (PGK) to illustrate enhanced sampling in BioEmu2 when an alternative reference conformation is known, but it is rarely covered by unsteered samples and its relative population under the model is not specified. 
For PGK, we define the bias potential using the relative pose of the two domains against the known closed conformation 2X13.
More details are described in Section~\ref{sec:steering-and-enhanced-sampling-example-pgk}.
In Fig. \ref{fig:steering}c, we compare unsteered sampling with enhanced sampling of the same total sample size.
The enhanced sampling preserves a quantitatively similar free-energy landscape in the region well sampled without steering and extends farther along the closed-reference-like arm.
Representative conformations illustrate the dominant upper-arm region (A), an open-reference-like conformation (B), and a closed-reference-like conformation (C). 

\section{Methods}
\label{sec:methods}

\subsection{Molecular System Definition}
\label{sec:methods-01-system-definition}

We describe a molecular system by
\begin{equation}
\mathcal{S}=(\mathcal{E},\mathcal{I},\mathcal{B}),
\end{equation}
where $\mathcal{E}$ specifies the molecular entities, $\mathcal{I}$
their chain copies, and $\mathcal{B}$ their covalent connectivity.
This separation distinguishes molecular identity from stoichiometry:
two identical protein chains share an entity definition but remain
distinct chains in the generated complex. The representation follows
the entity-and-chain organization of PDBx/mmCIF and the all-atom
modeling approach of AlphaFold 3 \citep{abramson2024alphafold3}.

\paragraph{Chemical composition and connectivity.}
\label{sec:methods-01-system-definition-chemical-composition-and-connectivity}
The system definition supports monomers, multichain assemblies, and
complexes containing proteins, RNA, DNA, modified residues, small
molecules and branched components. Proteins are specified by their
amino acid sequences, and RNA and DNA by their base sequences.
Chemical components can be identified by a Chemical Component
Dictionary (CCD) code \citep{westbrook2015ccd}, which specifies atom identities, bonds and
stereochemistry, or by supported text-based chemical encodings.
Ligands, including lipids and ions, accept a CCD code, SMILES or InChI;
SMILES and InChI describe the chemical structure directly.

Modified and noncanonical polymer residues are specified by their
sequence positions and a CCD code or InChI. This replaces the chemical
identity at the specified position.
For example, phosphothreonine can be specified by the CCD
code TPO at a threonine position. 

Standard polymer links are constructed from sequence order. Additional
covalent bonds can be added to specify connectivity not determined by chain order,
such as disulfides or covalent links between chains.
Together, the chain definitions, chain copies
and covalent bonds determine the molecular topology. Branched
components are representable but absent from the current training
data.

\paragraph{Atom- and token-level representations.}
\label{sec:methods-01-system-definition-atom--and-token-level-representations}
The system is expanded into an atom-level graph and a token sequence,
following the mixed-resolution representation used in all-atom
structure prediction \citep{abramson2024alphafold3}. Conventional polymer residues each occupy
one token; ligands, modified residues, and branched monomers use one
token per heavy atom. Thus, tokens are computational units rather than
a uniform coarse-grained description of the generated structure.

\subsection{Multiple Sequence Alignments}
\label{sec:msa}

Protein multiple sequence alignments (MSAs) provide evolutionary
information complementary to molecular composition. Sequence
conservation and correlated substitutions help the model identify
structural constraints within proteins and across interacting chains,
as in MSA-based structure-prediction methods \citep{mirdita2022colabfold,jumper2021alphafold}. BioEmu2 uses these
inputs for both structure prediction and equilibrium emulation.

\paragraph{Chain alignments and pairing.}
\label{sec:msa-chain-alignments-and-pairing}
An independent alignment is obtained for each distinct protein sequence. The default acquisition uses
ColabFold \citep{mirdita2022colabfold} to search for MSAs.
Each alignment includes the query sequence and homologous
sequences aligned to its residue positions. Non-protein components do
not require evolutionary alignments.

For heteromeric systems, paired alignments additionally associate
homologues across protein chains, for example by matching occurrences
of the two sequences within the genome of a single organism.
Paired rows supply a correspondence
that can inform intermolecular interactions, whilst unpaired alignments
provide intrachain evolutionary context without indicating a special relationship
between the separate alignments of the two chains.
BioEmu2 combines both kinds of rows and records which
positions originate from paired alignments. 

Schematically, alignment preparation yields
\begin{equation}
\{\text{chain alignments}\}
\longrightarrow
(A,d,q)
\longrightarrow
\text{MSA features},
\end{equation}
where $A_{ri}$ is the aligned symbol in row $r$ at token position $i$,
$d_{ri}$ records insertion information relative to the query, and
$q_{ri}$ is a binary flag that indicates pairing. Gaps represent absent aligned positions and chains with no sequence in a given row.
The concatenation of these features gives MSA feature vectors $u_{ri}$.
A coarse summary of the MSA information is also included in the trunk via a simple
mean-pooling over the MSA rows.

\subsection{Sequence Trunk and Conditioning}
\label{sec:methods-03-sequence-trunk-sequence-trunk-and-conditioning}
\label{sec:methods-03-sequence-trunk}

The sequence trunk converts chemical and evolutionary inputs into
trunk embeddings
$E_{\mathrm{trunk}}=(E_{\mathrm{single}},E_{\mathrm{pair}})$.
For $L$ tokens, the single-token and token-pair embeddings have shapes
\begin{equation}
E_{\mathrm{single}}\in\mathbb{R}^{L\times d_s},
\qquad
E_{\mathrm{pair}}\in\mathbb{R}^{L\times L\times d_z},
\end{equation}
where $d_s$ and $d_z$ are the single-token and token-pair feature
dimensions, respectively.
We set $d_s=384,d_z=256$ in the BioEmu2 model.
The trunk's outputs depend
on the molecular system, alignment, and conditioning. The conditioning inputs $\mathcal{C}_\text{trunk}$ consist of a mode label $m \in \{\text{equilibrium, structure prediction}\}$ and an optional set, $\mathcal{H}$, of contacts. The trunk is then evaluated as:
\begin{equation}
(E_{\mathrm{single}},E_{\mathrm{pair}})
=F_\theta(\mathcal{S},\mathrm{MSA},m,\mathcal{H}).
\end{equation}
This separation follows the representation-and-diffusion architecture
of AlphaFold 3 and Boltz \citep{abramson2024alphafold3,wohlwend2024boltz1}. It permits one representation
calculation to support many coordinate samples with unchanged
conditioning.

We use a PairMixer sequence trunk \citep{ouyangzhang2025pairmixer} that forgoes the triangular
attention and single-representation updates of the PairFormer \citep{abramson2024alphafold3}.
Earlier work, including Minifold \citep{wohlwend2025minifold}, also showed that a simpler trunk
architecture can achieve essentially the same performance as the
PairFormer.

\begin{figure}[tbp]
\centering
\includegraphics[width=\linewidth]{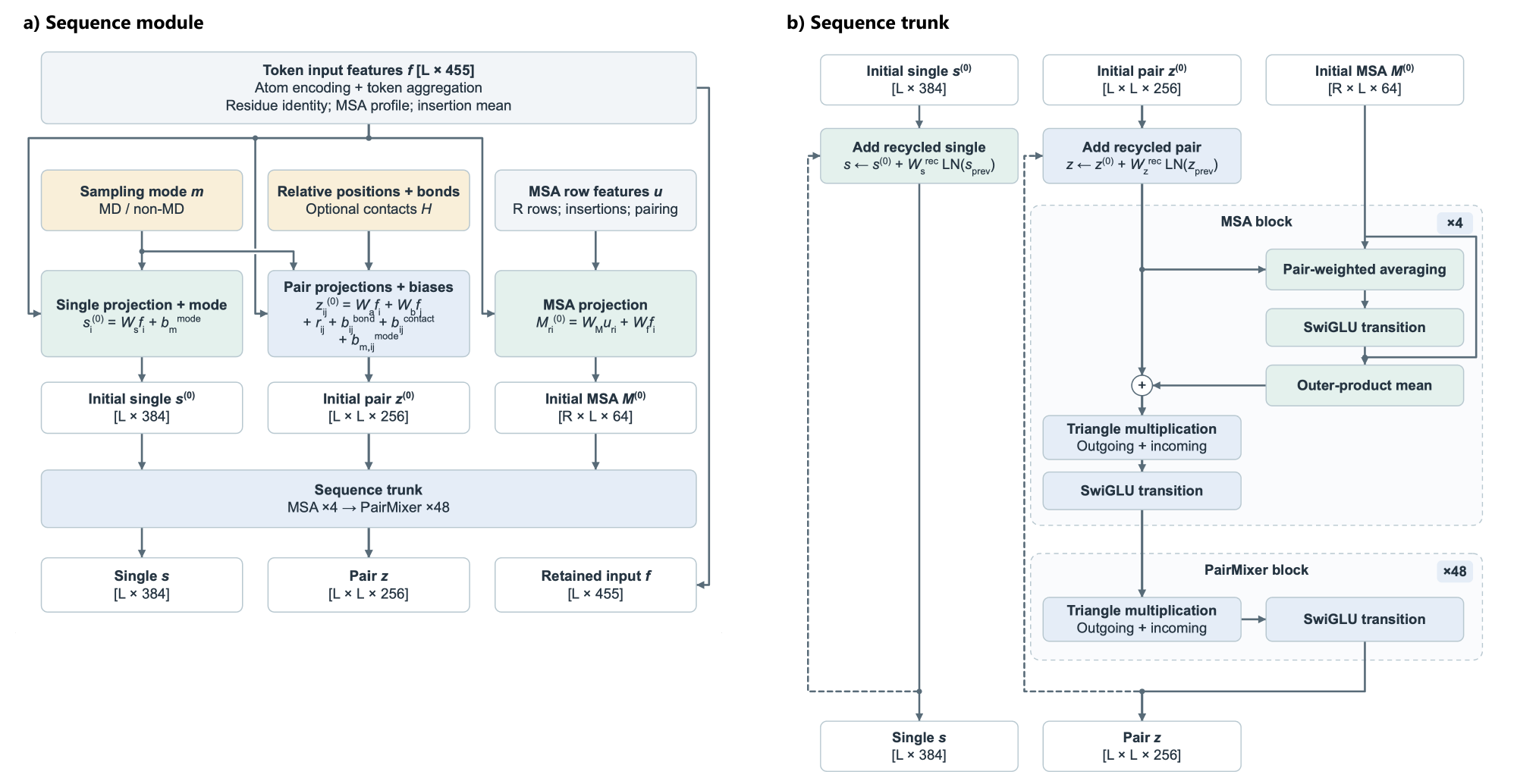}
\caption{\textbf{a) Sequence module.} The shared token input $f$ is projected into the
initial single and pair representations $s^{(0)}$ and $z^{(0)}$, and
contributes to the initial MSA representation $M^{(0)}$. The same
variables enter the trunk diagram below.
\textbf{b) Sequence trunk.} MSA processing and pair mixing refine the pair
representation, with optional recycling of single and pair states.
The single width is 384, the
pair width 256, and the token-input width 455.}
\label{fig:bioemu2-sequence-module}
\end{figure}

\paragraph{Initial features.}
\label{sec:methods-03-sequence-trunk-system-embedding}
Initial atom features for the trunk encode the charge, element, name and conformer reference position (giving a canonical geometry of individual residues, nucleotides and small molecules in isolation).
These features are then processed and pooled to the token level.
Additional initial token features are added to represent residue or nucleotide identity (for protein and nucleic acid chains), as well as coarse summaries of the MSA features (for protein chains).
The token input feature width of 455 is the concatenation
of a 384-dimensional aggregated atom encoding, a 33-dimensional
residue-type encoding, a 34-dimensional coarse MSA feature vector, and four un-used channels.

Given this feature vector $f_i$, the single and pair representations are initialized as
\begin{equation}
\begin{aligned}
s_i^{(0)}&=W_s f_i,\\
z_{ij}^{(0)}
&=W_a f_i+W_b f_j+r_{ij}+b_{ij}^{\mathrm{bond}},
\end{aligned}
\end{equation}
where $r_{ij}$ encodes information about the relative positions of $i, j$ such as whether they belong to the same chain or entity and their sequence separation and $b^\text{bond}_{ij}$ carries covalent bond information.

\paragraph{Mode conditioning.}
\label{sec:methods-03-sequence-trunk-mode-conditioning}
The sampling mode $m$ distinguishes MD-conditioned equilibrium
emulation from non-MD structure prediction. We implement this as a learned bias vector that is added to each token's initial single representation and a second bias that is added to the initial pair representation.

\paragraph{Contact conditioning.}
\label{sec:methods-03-sequence-trunk-contact-conditioning}
Following Boltz2~\cite{passaro2025boltz2}, the contact input is a list of token pairs and distance thresholds,
\begin{equation}
\mathcal{H}=\{(i,j,d_{ij}^{\max})\}.
\end{equation}
Each supplied pair is expected to lie within its threshold. 
The list
is converted into pairwise features: a category distinguishes selected
contacts, unselected pairs, and absence of contact conditioning. The
learned contact bias added to the pair representation is denoted
$b_{ij}^{\mathrm{contact}}$. For a selected pair,
\begin{equation}
b_{ij}^{\mathrm{contact}}
=\operatorname{Embed}_{\mathrm{contact}}
(\text{category}_{ij},d_{ij}^{\max}),
\end{equation}
the pair $(j, i)$ is updated symmetrically.
Contact conditioning in the trunk is distinct from steering, the latter being an inference-time means of influencing the sampling process.

\paragraph{MSA input.}
\label{sec:methods-03-sequence-trunk-msa-input}
Initial MSA features for row $r$ at token $i$ contain a one-hot encoding of the residue (or gap), and flags for insertion, deletion and pairing.
Given these MSA features $u_{ri}$, we combine them with initial token features $f_i$ via
\begin{equation}
M_{ri}^{(0)}=W_Mu_{ri}+W_ff_i.
\end{equation}

\paragraph{Trunk layers.}
\label{sec:methods-03-sequence-trunk-msa-processing-pair-mixing-and-recycling}
The trunk uses an MSA module followed by a PairMixer stack \citep{ouyangzhang2025pairmixer}.
The MSA module couples sequence rows through pair-conditioned
averaging and transfers alignment information to pairs through
outer-product aggregation.
After 4 MSA layers, the subsequent 48 PairMixer layers operate only on the pair representation.

Both the MSA module and PairMixer use
triangle multiplication, combining information about pairs that
share an intermediate token.
Suppressing learned projections, gates, and normalization, the two
triangle operations have the form
\begin{equation}
\Delta z_{ij}^{\mathrm{out}}\sim
\sum_k A(z_{ik})\odot B(z_{jk}),
\qquad
\Delta z_{ij}^{\mathrm{in}}\sim
\sum_k A'(z_{ki})\odot B'(z_{kj}),
\end{equation}
where $\odot$ denotes element-wise multiplication. These operations
communicate through triplets of tokens, allowing local chemical and
evolutionary evidence to inform relationships across the system.
PairMixer combines outgoing and incoming triangle updates with
pair-wise transition networks, without triangle-attention layers \citep{ouyangzhang2025pairmixer}.
The main stack refines pair features; single-token features retain
their input- and mode-dependent information.

\paragraph{Recycling.} We repeat the MSA and pair-processing stages while feeding
normalized, projected single and pair representations back into their
initial states. We use 3 recycling steps by default at inference time.
At training time, we use a deterministic schedule over recycling steps that keeps
all ranks in sync.

\subsection{Structure Module and Sampling}
\label{sec:structure-module-and-sampling}

\paragraph{Dual-stream structure architecture.}
\label{sec:structure-module-and-sampling-dual-stream-structure-architecture}
Our structure module is closely inspired by the Flow Blocks introduced
by NeuralPLexer3 \citep{qiao2026neuralplexer3}. Each of its 16 blocks interleaves atom-level and
token-level updates, rather than separating atom encoding, a token
transformer, and atom decoding as in AlphaFold 3 \citep{abramson2024alphafold3}.

The atom stream uses windowed attention and transition MLPs; the token
stream uses full token attention and transition MLPs. Cross-stream
updates gather features at designated token-center atoms and broadcast
token features to their associated atoms. Both streams are conditioned
on the trunk representations and diffusion noise level. For time
propagation, lag-time features and the supplied starting coordinates
provide additional conditioning. A coordinate readout follows the
stack, and EDM preconditioning combines this update with the noisy
input to obtain a denoised coordinate estimate.

\begin{figure}[tbp]
\centering
\includegraphics[width=0.6\linewidth]{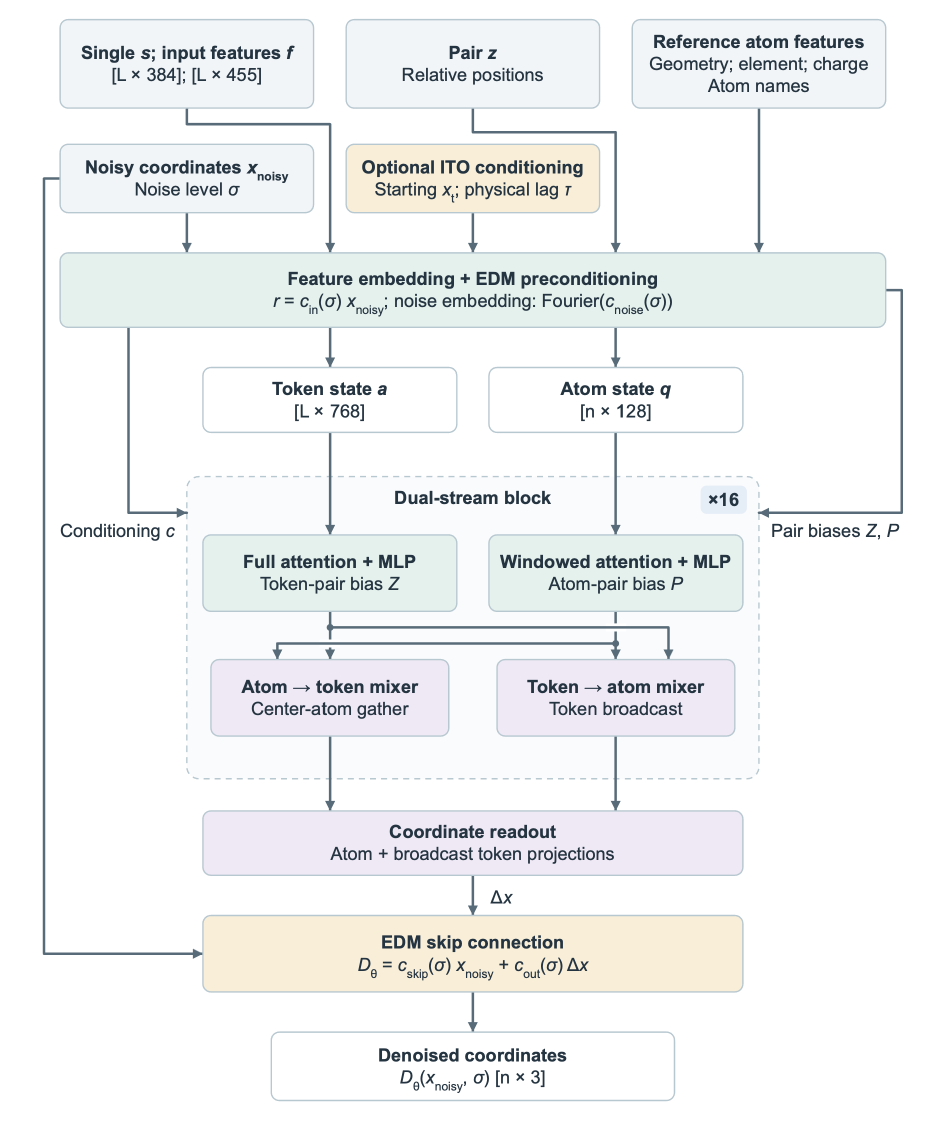}
\caption{\textbf{Structure module.} The output is the evaluated denoiser,
$\widehat{x}=D_\theta(x_\text{noisy},\sigma\mid\mathcal{C})$. Repeated sampling
updates use this prediction to generate coordinates.}
\label{fig:bioemu2-structure-module}
\end{figure}

\paragraph{Conditional denoising.}
\label{sec:structure-module-and-sampling-conditional-denoising}
After the biomolecular system has been embedded by the sequence trunk,
BioEmu2's structure model generates structures using a conditional
diffusion model. The conditions are defined as follows:
\begin{equation}
\mathcal{C}=
\left(\mathcal{S},E_{\mathrm{single}},E_{\mathrm{pair}};
[x_t,\tau]\right),
\end{equation}

\begin{itemize}
\item $\mathcal{S}$: the static system features described under
system embedding.
\item $E_{\mathrm{single}}$ and $E_{\mathrm{pair}}$: the single-token and
token-pair embeddings produced by the
sequence trunk.
\item $[x_t,\tau]$: the optional configuration at an initial physical time
and the positive lag $\tau$ for time propagation. When supplied
with equilibrium-mode conditioning, these specify sampling of a
future endpoint $x_{t+\tau}$ conditional on $x_t$.
\end{itemize}

The diffusion model is a conditional score model over all-atom Cartesian
coordinates. At each denoising step, it receives the current noisy
coordinates $x_\text{noisy}$ and their noise level $\sigma$, in addition
to $\mathcal{C}$. The static system features and embeddings are cached
across steps. The score estimates the gradient of the logarithm of the
noise-perturbed conditional density,
\begin{equation}
s_\theta(x_\text{noisy},\sigma\mid\mathcal{C})
\approx
\nabla_{x_\text{noisy}}
\log p_\sigma(x_\text{noisy}\mid\mathcal{C}),
\end{equation}
where $p_\sigma$ is the target structural distribution convolved with
Gaussian noise of covariance $\sigma^2 I$. BioEmu2 uses a denoiser
parameterization \citep{karras2022edm}: the network reconstructs clean coordinates
$D_\theta(x_\text{noisy},\sigma\mid\mathcal{C})$, which define the score
through
\begin{equation}
s_\theta(x_\text{noisy},\sigma\mid\mathcal{C})
=\frac{D_\theta(x_\text{noisy},\sigma\mid\mathcal{C})
-x_\text{noisy}}{\sigma^2}.
\end{equation}

Sampling from the model means progressively denoising the structure.
It starts from Gaussian coordinates
$x_{\text{noisy},\sigma_{\max}}\sim
\mathcal{N}(0,\sigma_{\max}^2 I)$ and progressively denoises them along a
decreasing noise schedule \citep{karras2022edm}. Without ITO conditioning, the
result is a sample from the selected structural distribution. With
$x_t$ and $\tau$, the result is a future configuration $x_{t+\tau}$.
(The noise level $\sigma$ and the sampler's denoising steps are distinct
from the physical lag $\tau$.)

\paragraph{Treatment of the ITO inputs.}
The time-conditioning inputs, $[x_t,\tau]$, enter only via the structure module. At training time, we apply noise $N(0,\varsigma^2)$ 
to the structure $x_t$ to improve the model's robustness to errors in the original structure ($\varsigma=0.1$\AA).
We featurize $\tau$ via $e_k(\tau)=e^{-\tau/s_k}\{1, \sin(2\pi q\tau/s_k),\cos(2\pi q\tau/s_k)\}$ concatenated over various decay rates $s_k$ and various $q$.
These features enter the single conditioning stream alongside the diffusion noise scale embedding. The time-conditioning coordinates $x_t$ enter alongside the noisy coordinates $x_\text{noisy}$, and are multiplied by a gate $g(\tau)$ which is constructed so that $g\to 0$ as $\tau \to \infty$.

\paragraph{EDM preconditioning and inference updates.}
\label{sec:structure-module-and-sampling-edm-preconditioning-and-inference-updates}
We follow the EDM diffusion model set-up \citep{karras2022edm}.
Writing $\Delta x_\theta$ for the network's coordinate-update readout, the
EDM denoiser \citep{karras2022edm} is
\begin{equation}
D_\theta(x_\text{noisy},\sigma\mid\mathcal{C})
=c_{\mathrm{skip}}(\sigma)x_\text{noisy}
+c_{\mathrm{out}}(\sigma)
\Delta x_\theta(c_{\mathrm{in}}(\sigma)x_\text{noisy},
c_{\mathrm{noise}}(\sigma)\mid\mathcal{C}),
\end{equation}
with
\begin{equation}
\begin{aligned}
c_{\mathrm{skip}}(\sigma)&=
\frac{\sigma_{\mathrm{data}}^2}
{\sigma^2+\sigma_{\mathrm{data}}^2},&
c_{\mathrm{out}}(\sigma)&=
\frac{\sigma\sigma_{\mathrm{data}}}
{\sqrt{\sigma^2+\sigma_{\mathrm{data}}^2}},\\
c_{\mathrm{in}}(\sigma)&=
\frac{1}{\sqrt{\sigma^2+\sigma_{\mathrm{data}}^2}},&
c_{\mathrm{noise}}(\sigma)&=
\frac{1}{4}\log\frac{\sigma}{\sigma_{\mathrm{data}}}.
\end{aligned}
\end{equation}
Here, $\sigma_{\mathrm{data}}$ sets the coordinate scale used for
preconditioning, 16\AA\ in our case. 

Let $\sigma_0>\cdots>\sigma_K=0$ be the inference noise schedule, and
let $x^{(0)}\sim\mathcal{N}(0,\sigma_0^2I)$. At step $k$, noise
injection (``churn'') temporarily increases the noise level,
\begin{equation}
\widehat{\sigma}_k=(1+\gamma_k)\sigma_k,
\qquad
x_{\text{noisy},k}=x^{(k)}
+S_{\mathrm{noise}}\sqrt{\widehat{\sigma}_k^2-\sigma_k^2}
\,\epsilon_k,
\qquad \epsilon_k\sim\mathcal{N}(0,I).
\end{equation}
We switch off churn below a certain noise level, simply setting $x_{\text{noisy},k}=x^{(k)}$ and $\widehat{\sigma}_k=\sigma_k$.
Given the structure $x_{\text{noisy}, k}$, one denoiser evaluation gives
$\widehat{x}_k=D_\theta(x_{\text{noisy},k},\widehat{\sigma}_k\mid\mathcal{C})$,
followed by
\begin{equation}
\label{eq:diffusion-step-update}
x_{\text{update},k}=\frac{x_{\text{noisy},k}-\widehat{x}_k}{\widehat{\sigma}_k},
\qquad
x^{(k+1)}=x_{\text{noisy},k}+
S_\text{step}(\sigma_{k+1}-\widehat{\sigma}_k)x_{\text{update},k}.
\end{equation}
Global rigid augmentations are applied
at each step. If an ITO conditioning structure $x_t$ is used, the same rigid augmentations are applied to it. At inference, as in AF3, the noisy coordinates at each reverse step are rigidly aligned to the denoised prediction before the update in \eqref{eq:diffusion-step-update}.

With direct steering, $\widehat{x}_k$ is replaced by a guided denoised
estimate before computing $x_{\text{update},k}$. The corresponding equations and
particle-correction variants are given in Sec.~\ref{sec:steering-and-enhanced-sampling}.

\paragraph{Structure-prediction sharpening.}
\label{sec:structure-module-and-sampling-structure-prediction-sharpening}
For structure prediction,
we apply empirical sharpening in the EDM sampling framework \citep{karras2022edm} by
amplifying the denoising updates relative to the injected noise. In the
update above, this corresponds to setting $S_\text{step}=1.5,S_\text{noise}=1.003$ in equation~\eqref{eq:diffusion-step-update}, compared with $S_\text{step}=S_\text{noise}=1$ for
equilibrium sampling. The structure prediction modification is also used in AlphaFold 3 \citep{abramson2024alphafold3} and
Boltz \citep{wohlwend2024boltz1,passaro2025boltz2}.

\subsection{Sampling the Equilibrium Distribution}
\label{sec:structure-module-and-sampling-sampling-the-equilibrium-distribution}

Our MD simulations use the isothermal-isobaric, or $NpT$, ensemble:
particle number $N$, external pressure $p$, and temperature $T$ are fixed,
while the simulation volume $V$ fluctuates. BioEmu2 targets equilibrium
sampling at ambient conditions: the $NpT$ training simulations used room
temperature $T$ and a pressure of $p=1$ atm. Let $x$ denote the
biomolecular coordinates generated by BioEmu2, and let $z$ collect the
remaining configurational degrees of freedom, including explicit
solvent. After integrating out momenta, the dimensionless reduced energy
of the full configuration is
\begin{equation}
u(x,z,V)=\beta\left[U(x,z;V)+pV\right],
\qquad
\beta=\frac{1}{k_{\mathrm B}T},
\end{equation}
where $U$ is the potential energy and $k_{\mathrm B}$ is Boltzmann's
constant. The joint equilibrium density is
\begin{equation}
\rho(x,z,V)=Z^{-1}\exp[-u(x,z,V)],
\end{equation}
with normalization constant $Z$ and the volume-dependent Cartesian
configuration domain understood. The equilibrium distribution relevant
to the model is the marginal over solvent and volume,
\begin{equation}
p_{eq}(x)=\int_0^\infty \mathrm{d}V
\int \mathrm{d}z\,\rho(x,z,V).
\end{equation}

BioEmu2 therefore includes solvent effects implicitly: it generates
biomolecular coordinates without explicit solvent coordinates, while
learning a distribution in which solvent and volume degrees of freedom
have been marginalized out. Thus, the structural distribution is not
simply the Boltzmann factor of an isolated protein.

BioEmu2 is a Boltzmann emulator which approximates this equilibrium distribution by
\begin{equation}
x\sim p_\theta,
\qquad
p_\theta(x)\approx p_{eq}(x),
\label{eq:equilibrium-limit}
\end{equation}
where $\theta$ denotes shared learned parameters and the dependence on
the system's sequence embeddings is suppressed for readability.
Equilibrium sampling is an approximation objective, not a guarantee of
calibrated populations or quantitative free energies for every system.

\subsection{Time Propagation}
\label{sec:structure-module-and-sampling-time-propagation-with-an-implicit-transfer-operator}
Markov state models (MSMs) describe molecular kinetics as
transitions between discrete conformational states at a chosen lag time $\tau$
\citep{schutte1999conformational,swope2004kinetics,prinz2011msm}. When many short trajectories collectively sample the relevant states and transitions, MSMs can recover long-time relaxation and equilibrium populations without requiring each
trajectory to equilibrate \citep{swope2004kinetics,prinz2011msm}.
Conventional MSMs, however, are system-specific analysis models and do not transfer across molecular systems.

More recent work has turned to generative neural networks, which can generalize across molecular systems. Examples include deep generative MSMs \citep{wu2018generativemsm}, Timewarp \citep{klein2023timewarp}, MDGen \citep{jing2024mdgen}, and transferable extensions of the implicit transfer operator (ITO) framework \citep{vigueradiez2026tito,antoniadis2026platito}. BioEmu2 follows the original ITO formulation \citep{schreiner2023ito}, in which a diffusion model conditioned on starting coordinates $x_t$ and a physical lag $\tau$ samples
\begin{equation}
x_{t+\tau}\sim p_{\theta,\tau}(\,\cdot\mid x_t),
\end{equation}
approximating the physical transition density $p_\tau(x_{t+\tau}\mid x_t)$. This formulation allows BioEmu2 to generate the endpoint $x_{t+\tau}$ directly, without resolving the intervening MD trajectory. Long-running, time-ordered trajectories can be sampled by repeated conditional sampling. Because only biomolecular coordinates are retained, other dynamical degrees of freedom, such as momenta and solvent, are generally not carried from one step to the next.

\paragraph{Equilibrium limit.} For ergodic systems, dynamics relax to a unique equilibrium distribution and the physical transition density loses memory of its initial configuration. For the model, the corresponding consistency condition is
\begin{equation}
\lim_{\tau\to\infty}p_{\theta,\tau}(x_{t+\tau}\mid x_t)=p_\theta(x_{t+\tau}),
\end{equation}
where $p_\theta$ is the equilibrium model as noted in Eq.~\eqref{eq:equilibrium-limit}.
We impose consistency with the empirical equilibrium defined by the molecular dynamics training data, rather than infer equilibrium from transitions as in MSMs.

To achieve this, we follow ideas from BoPITO \citep{vigueradiez2025bopito} and separate the transition density into a stationary part, given by the equilibrium model, and a time-dependent part that decays with the lag. In practice, the network does not add separate equilibrium and dynamic scores.
Instead, it removes the time-dependent part by fading out its dynamics conditioning as the lag grows. 
The lag-time features carry exponential envelopes
$e_k(\tau)$ and conditioning on the initial configuration is gated by a learned positive combination of these envelopes:
\begin{equation}
e_k(\tau)=\exp(-\tau/s_k),
\qquad
g(\tau)=\sum_k w_k e_k(\tau),
\qquad
w_k>0,\quad 0<s_k\le s_{\max}.
\end{equation}
The lag-time embedding added to the single representation takes the form of a damped harmonic oscillator and decays on the same timescales. With $s_{\max}=1$~ms, both the gate and the lag-time embedding vanish for $\tau\gg s_{\max}$, so only the stationary part remains and the network recovers the equilibrium denoiser exactly. The form of this crossover is an architectural choice, but the weights $w_k$ and timescales $s_k$ are learned from data. The crossover should therefore not be read as an estimate of a physical relaxation time: It guarantees consistency with the learned equilibrium distribution, which need not coincide with exact thermodynamic equilibrium.

\paragraph{Training.} For a fraction of the training samples, we draw pairs of frames $(x_t,x_{t+\tau})$ from the same MD trajectory. We rigidly align the target frame to the starting frame, then apply the same random rotation and translation to both, which preserves the motion relative to the starting configuration. The remaining MD frames serve as independent equilibrium examples, with dynamics conditioning masked out and the lag set to $\tau=\infty$. Both tasks share the same network and EDM denoising loss (Sec.~\ref{sec:structure-module-and-sampling-conditional-denoising}).

Nearly all of BioEmu2's dynamics training data comes from MD simulations of up to $1\,\mu\mathrm{s}$. Rates and pathways predicted for slower processes, such as binding, dissociation, folding, unfolding and other slow rearrangements, therefore require independent validation. Until the model is trained and validated on experimental data at these timescales, such predictions should be treated as qualitative hypotheses rather than quantitative kinetics.

\paragraph{Rollouts.} Each ITO step runs the full diffusion sampler, with every denoiser evaluation conditioned on the preceding configuration $x_t$ and the chosen lag $\tau$, while the trunk embeddings are computed once per system and reused across steps. With fixed sampler settings, the cost per step is independent of $\tau$, and the required compute grows linearly with the simulation time. The lag can vary between steps, and starting configurations can be experimental or predicted structures, equilibrium samples, or manually modified structures, such as an apo protein with ligands placed next to it.

\subsection{Steering and Enhanced Sampling}
\label{sec:steering-and-enhanced-sampling}

Steering changes the structural distribution sampled at inference time
without retraining the model. Unlike mode or contact conditioning,
which changes the trunk embeddings, it acts on coordinate generation.
It can favour physical geometry, particular conformations, or regions
that are rare under the unsteered model. BioEmu2 implements
potential-gradient guidance and particle-based Feynman-Kac methods
\citep{singhal2025fk,skreta2025fkc,wu2023practical}, which can also support enhanced-sampling workflows \citep{xie2026enhanceddiffusion}.

\paragraph{Potentials and the tilted distribution.}
\label{sec:steering-and-enhanced-sampling-potentials-and-the-tilted-distribution}
Let $p_{\mathrm{base}}(x)$ denote the unsteered distribution for fixed
system and conditioning. A dimensionless potential $v(x)$ specifies
the desired tilt,
\begin{equation}
p_v(x)=Z_v^{-1}p_{\mathrm{base}}(x)\exp[-v(x)],
\qquad
Z_v=\int p_{\mathrm{base}}(x)\exp[-v(x)]\,\mathrm{d}x.
\end{equation}
Lower potential values are favored. For a physical bias energy $B(x)$,
the corresponding dimensionless potential is $v(x)=\beta B(x)$;
geometric penalties can instead be defined directly in dimensionless
form. Multiple objectives can be combined by summing their potentials.
The tilt defines a target distribution, not a guarantee that every
guidance procedure samples it exactly.

\paragraph{Coordinate notation.} Following the notation of Section \ref{sec:structure-module-and-sampling}, $i$ indexes denoising steps rather than physical time.
The coordinates $x^{(i)}$ are the current noisy state at scheduled noise level $\sigma_i$; $x_{\text{noisy},i}$ denotes the coordinates after augmentation and noise injection, at reheated noise level $\hat\sigma_i$; and $\hat{x}_i=D_\theta(x_{\text{noisy},i},\hat\sigma_i\mid C)$ is the corresponding prediction of clean coordinates with conditioning $C$. We omit the particle index. The two procedures use these same coordinates but differ in their reheating schedules and in where the gradient is taken.

\paragraph{Physicality steering: geometry penalties.} Physicality potentials discourage local geometric defects during coordinate generation. We adapt the Boltz-steering potentials \citep{wohlwend2024boltz1} for ligand distance geometry, chirality, bond stereochemistry, planarity, and chain overlap, and extend them with explicit bond-length and bond-angle restraints and polymer clash penalties. 
The polymer clash penalties discourage atomic clashes within and between chains. The general intra-chain and inter-chain clash penalties exclude atom pairs connected within three covalent bonds. A separate local penalty prevents excessively close contacts between eligible protein or nucleic-acid atoms in different residues that are connected by two or three bonds.
The geometry restraints use tabulated reference values, including peptide and nucleotide linkages and declared disulfide bonds, with broad bounds where a matching reference is unavailable.

Following Boltz-steering, admissible intervals can be enforced with a flat-bottom penalty. A one-sided constraint omits the corresponding bound term. The violation vanishes within the allowed interval and grows linearly outside it; potential-specific scaling and sampler coefficients determine its influence. For a scalar geometric quantity $q$ with lower and upper bounds $q_{\min}$ and $q_{\max}$, the unweighted violation is
\begin{equation}
    h(q;q_{\min},q_{\max})=\max(q_{\min}-q,0)+\max(q-q_{\max},0).
\end{equation}
Other potentials can favour a specified contact, reference conformation,
or collective-variable range. 

\paragraph{Physicality steering: detached-coordinate guidance.}
\label{sec:steering-and-enhanced-sampling-physicality-steering}
We use the churn schedule of Section \ref{sec:structure-module-and-sampling} and evaluate $\hat{x}_i$ without differentiating through the denoiser. At each guided diffusion step, we initialize a separate optimization variable $y_i^{(0)}=\hat{x}_i$. Here $\ell$ counts inner potential-gradient steps, not diffusion steps, and $j$ indexes the potentials:
\begin{equation}
y_i^{(\ell+1)}=y_i^{(\ell)}-\sum_j\eta_{j,i,\ell}\nabla_y U_j(y_i^{(\ell)}),
\qquad
\hat x_i^{\,g}=y_i^{(L_i)}.
\end{equation}
After $L_i$ inner steps, $\hat{x}_i^{\,g}=y_i^{(L_i)}$ is the guided clean-coordinate prediction; the superscript $g$ denotes guidance. 
The coefficients $\eta_{j,i,\ell}$ include each potential's guidance strength and activation schedule, and guidance is applied on nonterminal transitions. 
We then apply the EDM update in Eq. \ref{eq:diffusion-step-update}, substituting the guided clean prediction $\hat{x}_i^g$ for $\hat{x}_i$. If resampling is enabled, each selected particle carries forward both its noisy coordinates and its guided prediction. If rigid alignment is enabled, we first align those noisy coordinates to that prediction.

Thus, the potential modifies the denoised prediction used for transport rather than backpropagating through the denoiser to its noisy input. Optional FK resampling uses differences of weighted potential energies evaluated before guidance, multiplied by a global weighting strength, together with a correction based on the carried guidance displacement and the current Gaussian noise. 

For the physicality comparison in Fig.~\ref{fig:steering}b, we used detached-coordinate gradient guidance without particle resampling, active during approximately the final 20\% of 200 denoising transitions. The sampling cohort and evaluation procedures are described in Section \ref{sec:method-benchmark-deets}.
Better local geometry does not guarantee better equilibrium populations, but helps to preserve and improve structure prediction metrics while significantly reduces unphysicalities.

\paragraph{Enhanced sampling: through-denoiser guidance and TDS weights.}
\label{sec:steering-and-enhanced-sampling-enhanced-sampling-tds}
For the PGK example, we instead use Twisted Diffusion Sampler (TDS) \cite{wu2023practical}. 
Let $v_i(x)$ be the bias on clean coordinates at step $i$, including any step-dependent scaling or changes to its parameters. The corresponding bias on noisy coordinates is $V_i(x_\text{noisy})=v_i(D_\theta(x_\text{noisy},\hat{\sigma}_i\mid C))$.
Here $V_i$ is a function of noisy input coordinates through the denoiser, unlike the physicality optimization over detached $y$. We use sigma reheating rather than the physicality churn schedule: $\hat\sigma_0=\sigma_0$ and $\hat\sigma_i=\sigma_{i-1}$ for $i>0$, with injected-noise variance $\nu_i=\hat\sigma_i^2-\sigma_i^2$. With $S_\text{step}=1$, the nonterminal proposal is
\begin{equation}
x^{(i+1)}=x_{\text{noisy},i}+\frac{\hat\sigma_i-\sigma_{i+1}}{\hat\sigma_i}
\left[D_\theta(x_{\text{noisy},i},\hat\sigma_i\mid C)-x_{\text{noisy},i}\right]-\frac{\nu_i}{2}\nabla_{x_{\text{noisy},i}}V_i(x_{\text{noisy},i}),
\label{eq:steering-x-update}
\end{equation}
The gradient acts on the noisy coordinates $x_{\text{noisy},i}$ through the denoiser. Let $w_i$ denote an unnormalized particle weight at reheated state $x_{\text{noisy},i}$. The log weights accumulate between resampling events according to
\begin{equation}
\log w_i=\log w_{i-1}-V_i(x_{\text{noisy},i})+V_{i-1}(x_{\text{noisy},{i-1}})+\log\frac{K_i}{Q_i},
\qquad i\geq1.
\label{eq:tds-weight}
\end{equation}
Here $K_i$ and $Q_i$ are the reference and guided-proposal transition densities evaluated at $x_{\text{noisy},i}$. They combine the preceding unguided or guided transport, respectively, with the current Gaussian refresh, conditional on the same sampled rigid augmentation. If their means differ by $\delta_i=m_{Q,i}-m_{K,i}$ and $r_i=x_{\text{noisy},i}-m_{Q,i}$ is the sampled Gaussian residual, then $\log(K_i/Q_i)=-r_i^\mathsf{T}\delta_i/\nu_i-\|\delta_i\|^2/(2\nu_i)$. The means are centered separately before applying the shared rotation and translation. 

The initial log weight is $-V_0(x_{\text{noisy},0})$, which need not vanish for constant-strength steering. 
On the last transition $m$, $\sigma_{m+1}=0$ and guidance is disabled. Setting the guidance term to zero in Eq.~\ref{eq:steering-x-update}, with $a=1$, gives the completed sample $X=x^{(m+1)}=D_\theta(x_{\text{noisy},m},\hat\sigma_m\mid C)$; the final log-weight increment is $-v(X)+V_m(x_{\text{noisy},m})$.
The resulting target is $p_v(X)\propto p_\theta(X)e^{-v(X)}$, where $p_\theta$ is the endpoint distribution of the finite-step unguided reference sampler.
Finite particle populations and weight degeneracy limit estimation of this target. 
Our PGK example used the discrete TDS kernel ratio in Eq. \ref{eq:tds-weight} \cite{wu2023practical}, not the Feynman–Kac Corrector (FKC) weight rule \cite{skreta2025fkc} as in the previous work \cite{xie2026enhanceddiffusion}. Unlike FKC, the TDS rule does not require $\partial_\sigma U$, a noise-level derivative of the potential through the denoiser.

\paragraph{Enhanced sampling: particle resampling.} We normalize weights within each particle group and compute $N_{\mathrm{eff}}=1/\sum_{n=1}^{N_p}\bar w_n^2$. The TDS implementation supports resampling at a configured interval or when any group's fraction $N_{\mathrm{eff}}/N_p$ falls below a threshold. Systematic resampling copies all ancestry-dependent state and resets the accumulated log weights; a final resampling is performed when potentials are supplied. The returned particles have equal weights but can share ancestors and should not be counted as independent observations.

\paragraph{Enhanced sampling: recovering unbiased model statistics.}
\label{sec:steering-and-enhanced-sampling-enhanced-sampling-and-recovery-of-model-statistics}

A bias potential can favor conformations that the unsteered model samples only rarely \cite{xie2026enhanceddiffusion}. 
In the tilted ensemble, the probability of a structure $x$ is proportional to its unsteered probability $p_\mathrm{base}(x)$ multiplied by $\exp[-v(x)]$. With sufficient overlap between the biased and unsteered ensembles, we can therefore estimate the unsteered mean of an observable $A$ by weighting each structure by the inverse factor:
\begin{equation}
\langle A\rangle_{\mathrm{base}}
=
\frac{\langle A(x)\exp[v(x)]\rangle_v}
{\langle\exp[v(x)]\rangle_v}.
\end{equation}

When multiple windows are used to apply different biases, their relative normalizations can be estimated with multistate reweighting using MBAR \citep{shirts2008mbar}. 
We remark that the steering and reweighting method can address rare-event sampling to recover statistics of the base model; 
they do not remove errors in the base model itself.

\paragraph{Example: PGK domain-pose enhanced sampling.} 
\label{sec:steering-and-enhanced-sampling-example-pgk}
To illustrate how a known alternative conformation can guide exploration without prescribing its population, 
we sampled apo PGK with equilibrium-mode conditioning and biased the relative translation and rotation of its N- and C-domain cores toward the closed reference 2X13. 
The results are shown in Fig. \ref{fig:steering}c.
We fitted the centered closed-reference cores to the corresponding sampled C-alpha coordinates, using 164 N-core and 186 C-core atoms. Let $R_N,R_C$ be the proper fitted rotations in row-vector convention, $c_N,c_C$ the sampled core centroids, and $c_N^\star,c_C^\star$ their closed-reference centroids. The bias collective variables are the translation error in the N-core frame and the relative-rotation chord error:
\begin{equation}
d(X)=\left\|(c_C-c_N)R_N^\mathsf{T}-(c_C^\star-c_N^\star)\right\|,
\qquad
c(X)=\frac{\left\|R_C R_N^\mathsf{T}-I\right\|_{\mathrm F}}{\sqrt{2}}.
\end{equation}

For a window with strength $k$, the bias $v(X;k)$ is quadratic outside the flat-bottom tolerances:
\begin{equation}
v(X;k)=\frac{k}{2}\left[
\left(\frac{[d(X)-d_0]_+}{w_d}\right)^2+
\left(\frac{[c(X)-c_0]_+}{w_c}\right)^2
\right],
\qquad [u]_+=\max(u,0).
\end{equation}

We used $d_0=1\,\mathrm{\AA}$, $w_d=2\,\mathrm{\AA}$, $c_0=2\sin(5^\circ/2)$, and $w_c=2\sin(15^\circ/2)$. This restrains the fitted interdomain pose rather than directly penalizing internal core RMSDs. 
We used two biased windows, $k=0.5$ and $k=1$, with TDS weights and 200 diffusion transitions per 14-particle population. Systematic resampling was triggered when $N_{\mathrm{eff}}/14<0.5$ and was also performed at the final endpoint. All 1,008 returned structures at each nonzero strength were retained without RMSD-based population screening.

For visualization, we computed independently rigid-aligned C-alpha RMSDs to open 2XE7 and closed 2X13 over a common 406-atom selection. 
The baseline panel (Fig.~\ref{fig:steering}c left) contains 4,032 unsteered structures. The enhanced panel (Fig.~\ref{fig:steering}c right) estimates the unbiased (i.e. $k=0$) distribution by ordinary MBAR \cite{shirts2008mbar}, pooling 2,016 baseline structures with the 1,008 structures from each steered ensemble with $k=0.5$ and $k=1$. 
We evaluated each saved structure under all three bias conditions, $k=0, 0.5, 1$, and supplied these bias energies and the sample counts to MBAR, which computes separate weights for estimating unbiased baseline statistics.
The free energy landscapes are rendered as contours of $\Delta F=-RT\log(p/p_{\max})$, with $T=300\,\mathrm K$.

The representative structures A-C were selected from saved source-pool frames in designated regions of this projection, subject to the protein geometry quality checks. These selections did not filter the landscapes. The structural illustrations use an N-core alignment and a common camera and scale, whereas the plotted RMSDs use independent whole-protein alignments to each reference.

\subsection{Benchmark Details}
\label{sec:method-benchmark-deets}

\paragraph{Structure prediction benchmarks.}
\label{sec:method-struct-benchmark}
We evaluated structure prediction using variations of FoldBench and Runs N’ Poses.
FoldBench covers protein and nucleic-acid monomers and six molecular-interface categories~\citep{xu2025benchmarking}; Runs N’ Poses evaluates protein–ligand co-folding across ligand–pocket similarity strata~\citep{vskrinjar2026evaluating}.

To ensure complete uniformity between competing methods, we decided to resample locally over the same task definition and MSAs, and then feed the samples to a common OpenStructure (OST) v2.11.1 pipeline \citep{studer2026fully}.
For every included model, we generated new predictions from the adapted tasks with the default prediction setup in publicly available code.
AlphaFold3~\citep{abramson2024alphafold3} was omitted for licensing reasons, and Boltz-2~\citep{passaro2025boltz2} because of its later training cut-off date.
ColabFold-based MSA variants were used where available~\citep{mirdita2022colabfold}; ESMFold2-Fast was evaluated without MSAs~\citep{Candido2026ESMFold2}.

FoldBench sampling tasks were converted via the AlphaFold3 pipeline following the authors' descriptions~\citep{xu2025benchmarking}.
Runs N’ Poses system definitions retained the upstream protein and ligand entities, using CCD definitions where possible and SMILES otherwise (e.g.~for glycan chains or peptides).
32 FoldBench systems are excluded due to common OST errors (14), common ligand modelling issues (8), problematic raw PDB deposition info (5), and DNA/RNA hybrids (5).
39 Runs N’ Poses systems were excluded owing to upstream fixes in PLINDER (21), token count exceeding 2,560 causing issues to multiple baselines (10), common OST errors (6), and common ligand modelling issues (2).
To avoid train--test pollution, we further removed 6 FoldBench and 584 Runs N’ Poses systems whose source PDB entries were deposited on or before 2021-09-30 but released afterward.
The final set after all exclusions mentioned above serves as the shared denominator for success rate.
Thus, unlike some other implementations of these benchmarks, we ensured that all models were compared on an identical set of test systems.

Comparisons used five generated structures per system.
We start with sampling five structures in the same batch on NVIDIA A100-80GB GPUs, and out-of-memory cases were retried with lower batch-size and/or rerun on B200 GPUs.

For a fair comparison without a confidence ranker, all models were assessed using best-of-five (oracle) selection.
Monomer performance was summarized by mean best-of-five lDDT.
Non-ligand interface success required DockQ $\ge0.23$; ligand success required RMSD $<2$ Å and lDDT-PLI $>0.8$ within a prediction.
Runs N’ Poses scored annotated proper ligands separately according to source assignments.
The upstream SuCOS-pocket similarity bins are used for binning and visualization, omitting ligands without pre-assigned numeric similarity. Uncertainty was estimated using 95\% percentile-bootstrap intervals from 1,000 target-level resamples.

\paragraph{Polymer geometry evaluation.} For Fig.~\ref{fig:steering}b, we compared five structures per sampling condition for each of 1,490 eligible systems. Accuracy uses the best-of-five criteria above; geometry includes all five structures, including those rejected by the saved filtering procedure. Both conditions were evaluated with identical OpenStructure 2.11.1 Scorer/Molck preparation and StereoCheck. A structure is counted as affected if at least one polymer clash, bond-length outlier, or bond-angle outlier is reported. Checks cover whole-system polymers and exclude ligand internal geometry and polymer-ligand clashes.

Geometry incidence is the affected fraction of the five structures from each unique system in a category, averaged across those systems. Categories may overlap. For this comparison, accuracy uses eligible interface or monomer target instances, with missing or unscorable targets counted as zero; these instances are the resampling units in the 1,000-replicate accuracy bootstrap. Geometry uses 5,000 paired system-bootstrap resamples, retaining all five structures and matching system identities between conditions, to obtain 95\% percentile intervals. The sampling campaigns differ in source revision, compilation, and numerical settings, so the comparison is descriptive rather than an isolated causal ablation of guidance.

\clearpage

\bibliographystyle{unsrtnat}
\bibliography{refs}

\begin{thebibliography}{72}
\providecommand{\natexlab}[1]{#1}
\providecommand{\url}[1]{\texttt{#1}}
\expandafter\ifx\csname urlstyle\endcsname\relax
  \providecommand{\doi}[1]{doi: #1}\else
  \providecommand{\doi}{doi: \begingroup \urlstyle{rm}\Url}\fi

\bibitem[Perutz(1970)]{perutz1970haemoglobin}
M.~F. Perutz.
\newblock Stereochemistry of cooperative effects in haemoglobin: Haem--haem interaction and the problem of allostery.
\newblock \emph{Nature}, 228:\penalty0 726--734, 1970.
\newblock \doi{10.1038/228726a0}.

\bibitem[Henzler-Wildman et~al.(2007)]{henzlerwildman2007intrinsic}
K.~A. Henzler-Wildman et~al.
\newblock Intrinsic motions along an enzymatic reaction trajectory.
\newblock \emph{Nature}, 450:\penalty0 838--844, 2007.
\newblock \doi{10.1038/nature06410}.

\bibitem[Metzker(2010)]{metzker2010sequencing}
M.~L. Metzker.
\newblock Sequencing technologies---the next generation.
\newblock \emph{Nature Reviews Genetics}, 11:\penalty0 31--46, 2010.
\newblock \doi{10.1038/nrg2626}.

\bibitem[Jumper et~al.(2021)]{jumper2021alphafold}
J.~Jumper et~al.
\newblock Highly accurate protein structure prediction with {AlphaFold}.
\newblock \emph{Nature}, 596:\penalty0 583--589, 2021.
\newblock \doi{10.1038/s41586-021-03819-2}.

\bibitem[Abramson et~al.(2024)]{abramson2024alphafold3}
J.~Abramson et~al.
\newblock Accurate structure prediction of biomolecular interactions with {AlphaFold 3}.
\newblock \emph{Nature}, 630:\penalty0 493--500, 2024.
\newblock \doi{10.1038/s41586-024-07487-w}.

\bibitem[Baek et~al.(2021)]{baek2021rosettafold}
M.~Baek et~al.
\newblock Accurate prediction of protein structures and interactions using a three-track neural network.
\newblock \emph{Science}, 373:\penalty0 871--876, 2021.
\newblock \doi{10.1126/science.abj8754}.

\bibitem[Krishna et~al.(2024)]{krishna2024rosettafoldallatom}
R.~Krishna et~al.
\newblock Generalized biomolecular modeling and design with {RoseTTAFold All-Atom}.
\newblock \emph{Science}, 384:\penalty0 eadl2528, 2024.
\newblock \doi{10.1126/science.adl2528}.

\bibitem[Bai et~al.(2015)Bai, McMullan, and Scheres]{bai2015cryoem}
X.-C. Bai, G.~McMullan, and S.~H.~W. Scheres.
\newblock How {cryo-EM} is revolutionizing structural biology.
\newblock \emph{Trends in Biochemical Sciences}, 40:\penalty0 49--57, 2015.
\newblock \doi{10.1016/j.tibs.2014.10.005}.

\bibitem[Kleckner and Foster(2011)]{kleckner2011nmrdynamics}
I.~R. Kleckner and M.~P. Foster.
\newblock An introduction to {NMR}-based approaches for measuring protein dynamics.
\newblock \emph{Biochimica et Biophysica Acta}, 1814:\penalty0 942--968, 2011.
\newblock \doi{10.1016/j.bbapap.2010.10.012}.

\bibitem[Lerner et~al.(2018)]{lerner2018smfret}
E.~Lerner et~al.
\newblock Toward dynamic structural biology: Two decades of single-molecule {F{\"o}rster} resonance energy transfer.
\newblock \emph{Science}, 359:\penalty0 eaan1133, 2018.
\newblock \doi{10.1126/science.aan1133}.

\bibitem[Ganim et~al.(2008)]{ganim2008amideir}
Z.~Ganim et~al.
\newblock Amide {I} two-dimensional infrared spectroscopy of proteins.
\newblock \emph{Accounts of Chemical Research}, 41:\penalty0 432--441, 2008.
\newblock \doi{10.1021/ar700188n}.

\bibitem[Svergun and Koch(2003)]{svergun2003saxs}
D.~I. Svergun and M.~H.~J. Koch.
\newblock Small-angle scattering studies of biological macromolecules in solution.
\newblock \emph{Reports on Progress in Physics}, 66:\penalty0 1735--1782, 2003.
\newblock \doi{10.1088/0034-4885/66/10/r05}.

\bibitem[Engen(2009)]{engen2009hdx}
J.~R. Engen.
\newblock Analysis of protein conformation and dynamics by hydrogen/deuterium exchange {MS}.
\newblock \emph{Analytical Chemistry}, 81:\penalty0 7870--7875, 2009.
\newblock \doi{10.1021/ac901154s}.

\bibitem[Markin et~al.(2021)Markin, Mokhtari, Sunden, Appel, Akiva, Longwell, Sabatti, Herschlag, and Fordyce]{markin2021htmek}
C.~J. Markin, D.~A. Mokhtari, F.~Sunden, M.~J. Appel, E.~Akiva, S.~A. Longwell, C.~Sabatti, D.~Herschlag, and P.~M. Fordyce.
\newblock Revealing enzyme functional architecture via high-throughput microfluidic enzyme kinetics.
\newblock \emph{Science}, 373:\penalty0 eabf8761, 2021.
\newblock \doi{10.1126/science.abf8761}.

\bibitem[Tsuboyama et~al.(2023)Tsuboyama, Dauparas, Chen, Laine, Mohseni~Behbahani, Weinstein, Mangan, Ovchinnikov, and Rocklin]{tsuboyama2023megascale}
K.~Tsuboyama, J.~Dauparas, J.~Chen, E.~Laine, Y.~Mohseni~Behbahani, J.~J. Weinstein, N.~M. Mangan, S.~Ovchinnikov, and G.~J. Rocklin.
\newblock Mega-scale experimental analysis of protein folding stability in biology and design.
\newblock \emph{Nature}, 620:\penalty0 434--444, 2023.
\newblock \doi{10.1038/s41586-023-06328-6}.

\bibitem[Lindorff-Larsen et~al.(2011)Lindorff-Larsen, Piana, Dror, and Shaw]{lindorfflarsen2011folding}
K.~Lindorff-Larsen, S.~Piana, R.~O. Dror, and D.~E. Shaw.
\newblock How fast-folding proteins fold.
\newblock \emph{Science}, 334:\penalty0 517--520, 2011.
\newblock \doi{10.1126/science.1208351}.

\bibitem[Plattner et~al.(2017)Plattner, Doerr, De~Fabritiis, and No{\'e}]{plattner2017association}
N.~Plattner, S.~Doerr, G.~De~Fabritiis, and F.~No{\'e}.
\newblock Complete protein-protein association kinetics in atomic detail revealed by molecular dynamics simulations and {Markov} modelling.
\newblock \emph{Nature Chemistry}, 9:\penalty0 1005--1011, 2017.
\newblock \doi{10.1038/nchem.2785}.

\bibitem[H{\'e}nin et~al.(2022)H{\'e}nin, Leli{\`e}vre, Shirts, Valsson, and Delemotte]{henin2022enhancedsampling}
J.~H{\'e}nin, T.~Leli{\`e}vre, M.~R. Shirts, O.~Valsson, and L.~Delemotte.
\newblock Enhanced sampling methods for molecular dynamics simulations.
\newblock \emph{Living Journal of Computational Molecular Science}, 4:\penalty0 1583, 2022.
\newblock \doi{10.33011/livecoms.4.1.1583}.

\bibitem[Wang et~al.(2019)]{wang2019cg}
J.~Wang et~al.
\newblock Machine learning of coarse-grained molecular dynamics force fields.
\newblock \emph{ACS Central Science}, 5:\penalty0 755--767, 2019.
\newblock \doi{10.1021/acscentsci.8b00913}.

\bibitem[Charron et~al.(2025)]{charron2025cg}
N.~E. Charron et~al.
\newblock Navigating protein landscapes with a machine-learned transferable coarse-grained model.
\newblock \emph{Nature Chemistry}, 17:\penalty0 1284--1292, 2025.
\newblock \doi{10.1038/s41557-025-01874-0}.

\bibitem[No{\'e} et~al.(2019)No{\'e}, Olsson, K{\"o}hler, and Wu]{noe2019boltzmanngenerators}
F.~No{\'e}, S.~Olsson, J.~K{\"o}hler, and H.~Wu.
\newblock Boltzmann generators: Sampling equilibrium states of many-body systems with deep learning.
\newblock \emph{Science}, 365:\penalty0 eaaw1147, 2019.
\newblock \doi{10.1126/science.aaw1147}.

\bibitem[Zheng et~al.(2024)]{zheng2024dig}
S.~Zheng et~al.
\newblock Predicting equilibrium distributions for molecular systems with deep learning.
\newblock \emph{Nature Machine Intelligence}, 6:\penalty0 558--567, 2024.
\newblock \doi{10.1038/s42256-024-00837-3}.

\bibitem[Jing et~al.(2024{\natexlab{a}})Jing, Berger, and Jaakkola]{jing2024alphaflow}
B.~Jing, B.~Berger, and T.~Jaakkola.
\newblock {AlphaFold} meets flow matching for generating protein ensembles.
\newblock In \emph{Proceedings of the 41st International Conference on Machine Learning}, volume 235 of \emph{Proceedings of Machine Learning Research}, pages 22277--22303. PMLR, 2024{\natexlab{a}}.
\newblock URL \url{https://proceedings.mlr.press/v235/jing24a.html}.

\bibitem[Lewis et~al.(2025)]{lewis2025bioemu}
S.~Lewis et~al.
\newblock Scalable emulation of protein equilibrium ensembles with generative deep learning.
\newblock \emph{Science}, 389:\penalty0 eadv9817, 2025.
\newblock \doi{10.1126/science.adv9817}.

\bibitem[Klein et~al.(2023)]{klein2023timewarp}
L.~Klein et~al.
\newblock {Timewarp}: Transferable acceleration of molecular dynamics by learning time-coarsened dynamics.
\newblock arXiv:2302.01170, 2023.
\newblock URL \url{https://arxiv.org/abs/2302.01170}.

\bibitem[Jing et~al.(2024{\natexlab{b}})Jing, St{\"a}rk, Jaakkola, and Berger]{jing2024mdgen}
B.~Jing, H.~St{\"a}rk, T.~Jaakkola, and B.~Berger.
\newblock Generative modeling of molecular dynamics trajectories.
\newblock In \emph{Advances in Neural Information Processing Systems}, volume~37, pages 40534--40564, 2024{\natexlab{b}}.
\newblock \doi{10.52202/079017-1282}.

\bibitem[Schreiner et~al.(2023)Schreiner, Winther, and Olsson]{schreiner2023ito}
M.~Schreiner, O.~Winther, and S.~Olsson.
\newblock Implicit transfer operator learning: Multiple time-resolution models for molecular dynamics.
\newblock In \emph{Advances in Neural Information Processing Systems}, volume~36, pages 36449--36462, 2023.
\newblock URL \url{https://arxiv.org/abs/2305.18046}.

\bibitem[Wohlwend et~al.(2024)]{wohlwend2024boltz1}
J.~Wohlwend et~al.
\newblock {Boltz-1}: Democratizing biomolecular interaction modeling.
\newblock \emph{bioRxiv}, 2024.
\newblock \doi{10.1101/2024.11.19.624167}.

\bibitem[Zhang et~al.(2026)Zhang, Gong, Sun, Guan, Ren, Xue, Zhang, Ma, Liu, Chen, et~al.]{zhang2026protenix}
Y.~Zhang, C.~Gong, J.~Sun, J.~Guan, M.~Ren, S.~Xue, H.~Zhang, W.~Ma, Z.~Liu, X.~Chen, et~al.
\newblock Protenix-v2: Broadening the reach of structure prediction and biomolecular design.
\newblock \emph{bioRxiv}, 2026.
\newblock \doi{10.64898/2026.04.10.717613}.

\bibitem[{Aureka AI OpenDDE project}(2026)]{opendde}
{Aureka AI OpenDDE project}.
\newblock Folding, reasoning, and scaling with open-source drug discovery engine.
\newblock arXiv:2607.03787, 2026.
\newblock URL \url{https://arxiv.org/abs/2607.03787}.

\bibitem[Candido et~al.(2026)Candido, Hayes, Derry, Rao, Lin, Verkuil, Wu, Lee, Bruguera, Keval, Kopylov, Pak, Wu, Thomas, Mataraso, Hsu, Trotman-Grant, Fatras, dos Santos~Costa, Badkundri, Akın, Oktay, Deaton, Montabana, Sitwala, Yu, Wiggert, Carlin, Goering, Blazejewski, Sandora, Hla, Jia, Kloker, Sofroniew, Uehara, Pannu, Bachas, Liu, Sercu, and Rives]{Candido2026ESMFold2}
S.~Candido, T.~Hayes, A.~Derry, R.~Rao, Z.~Lin, R.~Verkuil, B.~Z. Wu, J.~S. Lee, E.~S. Bruguera, J.~A. Keval, M.~Kopylov, J.~E. Pak, W.~Wu, N.~Thomas, S.~Mataraso, A.~Hsu, A.~C. Trotman-Grant, K.~Fatras, A.~dos Santos~Costa, R.~Badkundri, H.~Akın, D.~Oktay, J.~Deaton, E.~Montabana, H.~Sitwala, Y.~Yu, M.~Wiggert, D.~A. Carlin, A.~W. Goering, T.~Blazejewski, M.~Sandora, M.~Hla, T.~Z. Jia, L.~H. Kloker, N.~J. Sofroniew, M.~Uehara, J.~Pannu, S.~Bachas, D.~S. Liu, T.~Sercu, and A.~Rives.
\newblock Language modeling materializes a world model of protein biology.
\newblock \emph{bioRxiv}, 2026.
\newblock \doi{10.64898/2026.06.03.729735}.

\bibitem[Xu et~al.(2026)Xu, Feng, Qiao, Wu, Shen, Cheng, Zheng, and Sun]{xu2025benchmarking}
S.~Xu, Q.~Feng, L.~Qiao, H.~Wu, T.~Shen, Y.~Cheng, S.~Zheng, and S.~Sun.
\newblock Benchmarking all-atom biomolecular structure prediction with foldbench.
\newblock \emph{Nature Communications}, 17:\penalty0 442, 2026.
\newblock \doi{10.1038/s41467-025-67127-3}.

\bibitem[{\v{S}}krinjar et~al.(2026){\v{S}}krinjar, Eberhardt, Studer, Tauriello, Schwede, and Durairaj]{vskrinjar2026evaluating}
P.~{\v{S}}krinjar, J.~Eberhardt, G.~Studer, G.~Tauriello, T.~Schwede, and J.~Durairaj.
\newblock Evaluating generalization in protein--ligand cofolding methods.
\newblock \emph{Nature Structural \& Molecular Biology}, 33:\penalty0 782--794, 2026.
\newblock \doi{10.1038/s41594-026-01797-5}.

\bibitem[Steinegger and S{\"o}ding(2017)]{steinegger2017mmseqs2}
M.~Steinegger and J.~S{\"o}ding.
\newblock {MMseqs2} enables sensitive protein sequence searching for the analysis of massive data sets.
\newblock \emph{Nature Biotechnology}, 35:\penalty0 1026--1028, 2017.
\newblock \doi{10.1038/nbt.3988}.

\bibitem[Mirdita et~al.(2022)Mirdita, Sch{\"u}tze, Moriwaki, Heo, Ovchinnikov, and Steinegger]{mirdita2022colabfold}
M.~Mirdita, K.~Sch{\"u}tze, Y.~Moriwaki, L.~Heo, S.~Ovchinnikov, and M.~Steinegger.
\newblock {ColabFold}: making protein folding accessible to all.
\newblock \emph{Nature Methods}, 19:\penalty0 679--682, 2022.
\newblock \doi{10.1038/s41592-022-01488-1}.

\bibitem[Sohl-Dickstein et~al.(2015)Sohl-Dickstein, Weiss, Maheswaranathan, and Ganguli]{sohldickstein2015deep}
J.~Sohl-Dickstein, E.~Weiss, N.~Maheswaranathan, and S.~Ganguli.
\newblock Deep unsupervised learning using nonequilibrium thermodynamics.
\newblock In \emph{Proceedings of the 32nd International Conference on Machine Learning}, volume~37 of \emph{Proceedings of Machine Learning Research}, pages 2256--2265. PMLR, 2015.
\newblock URL \url{https://proceedings.mlr.press/v37/sohl-dickstein15.html}.

\bibitem[Ho et~al.(2020)Ho, Jain, and Abbeel]{ho2020denoising}
J.~Ho, A.~Jain, and P.~Abbeel.
\newblock Denoising diffusion probabilistic models.
\newblock In \emph{Advances in Neural Information Processing Systems}, volume~33, pages 6840--6851, 2020.
\newblock URL \url{https://proceedings.neurips.cc/paper/2020/hash/4c5bcfec8584af0d967f1ab10179ca4b-Abstract.html}.

\bibitem[Song et~al.(2021)Song, Sohl-Dickstein, Kingma, Kumar, Ermon, and Poole]{song2021scorebased}
Y.~Song, J.~Sohl-Dickstein, D.~P. Kingma, A.~Kumar, S.~Ermon, and B.~Poole.
\newblock Score-based generative modeling through stochastic differential equations.
\newblock In \emph{International Conference on Learning Representations}, 2021.
\newblock URL \url{https://arxiv.org/abs/2011.13456}.

\bibitem[Passaro et~al.(2025)]{passaro2025boltz2}
S.~Passaro et~al.
\newblock {Boltz-2}: Towards accurate and efficient binding affinity prediction.
\newblock \emph{bioRxiv}, 2025.
\newblock \doi{10.1101/2025.06.14.659707}.

\bibitem[{The OpenFold3 Team}(2025)]{openfold3-preview}
{The OpenFold3 Team}.
\newblock Openfold3-preview, 2025.
\newblock URL \url{https://github.com/aqlaboratory/openfold-3}.

\bibitem[Mariani et~al.(2013)Mariani, Biasini, Barbato, and Schwede]{mariani2013lddt}
V.~Mariani, M.~Biasini, A.~Barbato, and T.~Schwede.
\newblock {lDDT}: a local superposition-free score for comparing protein structures and models using distance difference tests.
\newblock \emph{Bioinformatics}, 29:\penalty0 2722--2728, 2013.
\newblock \doi{10.1093/bioinformatics/btt473}.

\bibitem[Basu and Wallner(2016)]{basu2016dockq}
S.~Basu and B.~Wallner.
\newblock {DockQ}: A quality measure for protein-protein docking models.
\newblock \emph{PLOS ONE}, 11:\penalty0 e0161879, 2016.
\newblock \doi{10.1371/journal.pone.0161879}.

\bibitem[Lin et~al.(2026)Lin, Chu, Iyer, Lee, St~John, Boyd, Roland, Ren, Zhou, Cao, et~al.]{lin2026fold}
D.~Lin, S.~Chu, V.~Iyer, Y.~Lee, J.~St~John, K.~Boyd, B.~Roland, X.~Ren, G.~Zhou, Z.~Cao, et~al.
\newblock Fold-cp: A context parallelism framework for biomolecular modeling.
\newblock arXiv:2603.14806, 2026.
\newblock URL \url{https://arxiv.org/abs/2603.14806}.

\bibitem[Wang et~al.(2004)Wang, Wolf, Caldwell, Kollman, and Case]{wang2004development}
J.~Wang, R.~M. Wolf, J.~W. Caldwell, P.~A. Kollman, and D.~A. Case.
\newblock Development and testing of a general amber force field.
\newblock \emph{Journal of Computational Chemistry}, 25:\penalty0 1157--1174, 2004.
\newblock \doi{10.1002/jcc.20035}.

\bibitem[P{\'e}rez-Hern{\'a}ndez et~al.(2013)P{\'e}rez-Hern{\'a}ndez, Paul, Giorgino, De~Fabritiis, and No{\'e}]{perezhernandez2013tica}
G.~P{\'e}rez-Hern{\'a}ndez, F.~Paul, T.~Giorgino, G.~De~Fabritiis, and F.~No{\'e}.
\newblock Identification of slow molecular order parameters for {Markov} model construction.
\newblock \emph{The Journal of Chemical Physics}, 139:\penalty0 015102, 2013.
\newblock \doi{10.1063/1.4811489}.

\bibitem[Schwantes and Pande(2013)]{schwantes2013tica}
C.~R. Schwantes and V.~S. Pande.
\newblock Improvements in {Markov} state model construction reveal many non-native interactions in the folding of {NTL9}.
\newblock \emph{Journal of Chemical Theory and Computation}, 9:\penalty0 2000--2009, 2013.
\newblock \doi{10.1021/ct300878a}.

\bibitem[Olsson et~al.(2026)Olsson, No{\'e}, Rotskoff, and Lindorff-Larsen]{olsson2026streetlight}
S.~Olsson, F.~No{\'e}, G.~M. Rotskoff, and K.~Lindorff-Larsen.
\newblock Looking under the streetlight: Evaluation in generative molecular dynamics.
\newblock In \emph{Advances in Neural Information Processing Systems}, 2026.
\newblock URL \url{https://arxiv.org/abs/2610.12518}.
\newblock Position Track.

\bibitem[Viguera~Diez et~al.(2026)Viguera~Diez, Schreiner, and Olsson]{vigueradiez2026tito}
J.~Viguera~Diez, M.~Schreiner, and S.~Olsson.
\newblock Transferable generative models bridge femtosecond to nanosecond time-step molecular dynamics.
\newblock \emph{Science Advances}, 12:\penalty0 eaed2333, 2026.
\newblock \doi{10.1126/sciadv.aed2333}.

\bibitem[Berry et~al.(1994)Berry, Meador, Bilderback, Liang, Glaser, and Phillips]{berry1994adk}
M.~B. Berry, B.~Meador, T.~Bilderback, P.~Liang, M.~Glaser, and G.~N. Phillips.
\newblock The closed conformation of a highly flexible protein: The structure of {\textit{e. coli}} adenylate kinase with bound {AMP} and {AMPPNP}.
\newblock \emph{Proteins: Structure, Function, and Bioinformatics}, 19:\penalty0 183--198, 1994.
\newblock \doi{10.1002/prot.340190304}.

\bibitem[Day et~al.(2008)Day, Smits, Fan, Lee, Fairlie, and Hinds]{day20082roc}
C.~L. Day, C.~Smits, F.~C. Fan, E.~F. Lee, W.~D. Fairlie, and M.~G. Hinds.
\newblock Structure of the bh3 domains from the p53-inducible bh3-only proteins noxa and puma in complex with mcl-1.
\newblock \emph{Journal of Molecular Biology}, 380:\penalty0 958--971, 2008.
\newblock \doi{10.1016/j.jmb.2008.05.071}.

\bibitem[Rogers et~al.(2013)Rogers, Steward, and Clarke]{rogers2013puma}
J.~M. Rogers, A.~Steward, and J.~Clarke.
\newblock Folding and binding of an intrinsically disordered protein: Fast, but not ‘diffusion-limited’.
\newblock \emph{Journal of the American Chemical Society}, 135:\penalty0 1415--1422, 2013.
\newblock \doi{10.1021/ja309527h}.

\bibitem[Xie et~al.(2026)]{xie2026enhanceddiffusion}
Y.~Xie et~al.
\newblock Enhanced diffusion sampling: Efficient rare event sampling and free energy calculation with diffusion models.
\newblock arXiv:2602.16634, 2026.
\newblock URL \url{https://arxiv.org/abs/2602.16634}.

\bibitem[Singhal et~al.(2025)]{singhal2025fk}
R.~Singhal et~al.
\newblock A general framework for inference-time scaling and steering of diffusion models.
\newblock In \emph{Proceedings of the 42nd International Conference on Machine Learning}, volume 267 of \emph{Proceedings of Machine Learning Research}, pages 55810--55827. PMLR, 2025.
\newblock URL \url{https://proceedings.mlr.press/v267/singhal25b.html}.

\bibitem[Wu et~al.(2023)Wu, Trippe, Naesseth, Blei, and Cunningham]{wu2023practical}
L.~Wu, B.~Trippe, C.~Naesseth, D.~Blei, and J.~Cunningham.
\newblock Practical and asymptotically exact conditional sampling in diffusion models.
\newblock In \emph{Advances in Neural Information Processing Systems}, volume~36, pages 31372--31403, 2023.
\newblock URL \url{https://papers.nips.cc/paper_files/paper/2023/hash/63e8bc7bbf1cfea36d1d1b6538aecce5-Abstract-Conference.html}.

\bibitem[Zwanzig(1954)]{zwanzig1954perturbation}
R.~W. Zwanzig.
\newblock High-temperature equation of state by a perturbation method. {I}. nonpolar gases.
\newblock \emph{The Journal of Chemical Physics}, 22:\penalty0 1420--1426, 1954.
\newblock \doi{10.1063/1.1740409}.

\bibitem[Torrie and Valleau(1977)]{torrie1977umbrella}
G.~M. Torrie and J.~P. Valleau.
\newblock Nonphysical sampling distributions in {Monte Carlo} free-energy estimation: Umbrella sampling.
\newblock \emph{Journal of Computational Physics}, 23:\penalty0 187--199, 1977.
\newblock \doi{10.1016/0021-9991(77)90121-8}.

\bibitem[Skreta et~al.(2025)]{skreta2025fkc}
M.~Skreta et~al.
\newblock Feynman-kac correctors in diffusion: Annealing, guidance, and product of experts.
\newblock In \emph{Proceedings of the 42nd International Conference on Machine Learning}, volume 267 of \emph{Proceedings of Machine Learning Research}, pages 55906--55949. PMLR, 2025.
\newblock URL \url{https://proceedings.mlr.press/v267/skreta25a.html}.

\bibitem[Bartels(2000)]{bartels2000biased}
C.~Bartels.
\newblock Analyzing biased {Monte Carlo} and molecular dynamics simulations.
\newblock \emph{Chemical Physics Letters}, 331:\penalty0 446--454, 2000.
\newblock \doi{10.1016/S0009-2614(00)01215-X}.

\bibitem[Kong et~al.(2003)Kong, McCullagh, Meng, Nicolae, and Tan]{kong2003theory}
A.~Kong, P.~McCullagh, X.-L. Meng, D.~Nicolae, and Z.~Tan.
\newblock A theory of statistical models for {Monte Carlo} integration.
\newblock \emph{Journal of the Royal Statistical Society: Series B (Statistical Methodology)}, 65:\penalty0 585--604, 2003.
\newblock \doi{10.1111/1467-9868.00404}.

\bibitem[Shirts and Chodera(2008)]{shirts2008mbar}
M.~R. Shirts and J.~D. Chodera.
\newblock Statistically optimal analysis of samples from multiple equilibrium states.
\newblock \emph{The Journal of Chemical Physics}, 129:\penalty0 124105, 2008.
\newblock \doi{10.1063/1.2978177}.

\bibitem[Westbrook et~al.(2015)]{westbrook2015ccd}
J.~D. Westbrook et~al.
\newblock The chemical component dictionary: complete descriptions of constituent molecules in experimentally determined {3D} macromolecules in the {Protein Data Bank}.
\newblock \emph{Bioinformatics}, 31:\penalty0 1274--1278, 2015.
\newblock \doi{10.1093/bioinformatics/btu789}.

\bibitem[Ouyang-Zhang et~al.(2025)]{ouyangzhang2025pairmixer}
J.~Ouyang-Zhang et~al.
\newblock Triangle multiplication is all you need for biomolecular structure representations.
\newblock arXiv:2510.18870, 2025.
\newblock URL \url{https://arxiv.org/abs/2510.18870}.

\bibitem[Wohlwend et~al.(2025)Wohlwend, Reveiz, McPartlon, Feldmann, Jin, and Barzilay]{wohlwend2025minifold}
J.~Wohlwend, M.~Reveiz, M.~McPartlon, A.~Feldmann, W.~Jin, and R.~Barzilay.
\newblock {Minifold}: Simple, fast, and accurate protein structure prediction.
\newblock \emph{Transactions on Machine Learning Research}, 2025.
\newblock URL \url{https://openreview.net/forum?id=1p9hQTbjgo}.

\bibitem[Qiao et~al.(2025)Qiao, Ding, Dresselhaus, Rosenfeld, Han, Howell, Iyengar, Opalenski, Christensen, Sirumalla, et~al.]{qiao2026neuralplexer3}
J.~Z. Qiao, F.~Ding, T.~Dresselhaus, M.~Rosenfeld, X.~Han, O.~Howell, A.~Iyengar, S.~Opalenski, A.~Christensen, S.~K. Sirumalla, et~al.
\newblock {NeuralPLexer3}: accurate biomolecular complex structure prediction with flow models.
\newblock In \emph{Advances in Neural Information Processing Systems}, volume~38, pages 1--38, 2025.
\newblock \doi{10.52202/085713-0001}.

\bibitem[Karras et~al.(2022)Karras, Aittala, Aila, and Laine]{karras2022edm}
T.~Karras, M.~Aittala, T.~Aila, and S.~Laine.
\newblock Elucidating the design space of diffusion-based generative models.
\newblock In \emph{Advances in Neural Information Processing Systems}, volume~35, pages 26565--26577, 2022.
\newblock URL \url{https://arxiv.org/abs/2206.00364}.

\bibitem[Sch{\"u}tte et~al.(1999)Sch{\"u}tte, Fischer, Huisinga, and Deuflhard]{schutte1999conformational}
C.~Sch{\"u}tte, A.~Fischer, W.~Huisinga, and P.~Deuflhard.
\newblock A direct approach to conformational dynamics based on hybrid monte carlo.
\newblock \emph{Journal of Computational Physics}, 151:\penalty0 146--168, 1999.
\newblock \doi{10.1006/jcph.1999.6231}.

\bibitem[Swope et~al.(2004)Swope, Pitera, and Suits]{swope2004kinetics}
W.~C. Swope, J.~W. Pitera, and F.~Suits.
\newblock Describing protein folding kinetics by molecular dynamics simulations. 1. theory.
\newblock \emph{The Journal of Physical Chemistry B}, 108:\penalty0 6571--6581, 2004.
\newblock \doi{10.1021/jp037421y}.

\bibitem[Prinz et~al.(2011)]{prinz2011msm}
J.-H. Prinz et~al.
\newblock Markov models of molecular kinetics: Generation and validation.
\newblock \emph{The Journal of Chemical Physics}, 134:\penalty0 174105, 2011.
\newblock \doi{10.1063/1.3565032}.

\bibitem[Wu et~al.(2018)Wu, Mardt, Pasquali, and No{\'e}]{wu2018generativemsm}
H.~Wu, A.~Mardt, L.~Pasquali, and F.~No{\'e}.
\newblock Deep generative markov state models.
\newblock In \emph{Advances in Neural Information Processing Systems}, volume~31, 2018.
\newblock URL \url{https://arxiv.org/abs/1805.07601}.

\bibitem[Antoniadis et~al.(2026)Antoniadis, Pavesi, Olsson, and Winther]{antoniadis2026platito}
P.~Antoniadis, B.~Pavesi, S.~Olsson, and O.~Winther.
\newblock Protein language model embeddings improve generalization of implicit transfer operators.
\newblock arXiv:2602.11216, 2026.
\newblock URL \url{https://arxiv.org/abs/2602.11216}.

\bibitem[Viguera~Diez et~al.(2025)Viguera~Diez, Schreiner, Engkvist, and Olsson]{vigueradiez2025bopito}
J.~Viguera~Diez, M.~Schreiner, O.~Engkvist, and S.~Olsson.
\newblock Boltzmann priors for implicit transfer operators.
\newblock In \emph{International Conference on Learning Representations}, 2025.
\newblock URL \url{https://arxiv.org/abs/2410.10605}.

\bibitem[Studer et~al.(2026)Studer, Robin, Bienert, Durairaj, {\v{S}}krinjar, Tauriello, Waterhouse, and Schwede]{studer2026fully}
G.~Studer, X.~Robin, S.~Bienert, J.~Durairaj, P.~{\v{S}}krinjar, G.~Tauriello, A.~M. Waterhouse, and T.~Schwede.
\newblock A fully automated benchmarking suite to compare macromolecular complexes.
\newblock \emph{Nature Methods}, 23:\penalty0 387--394, 2026.
\newblock \doi{10.1038/s41592-025-02973-z}.

\end{thebibliography}

\end{document}